\documentclass[11pt]{article}

\usepackage[final]{acl}
\usepackage{amsfonts}
\usepackage{times}
\usepackage{amsmath}
\usepackage{latexsym}
\usepackage{subcaption}
\usepackage{algorithm}
\usepackage{algpseudocode}
\usepackage{pifont}
\usepackage[T1]{fontenc}
\usepackage{placeins}
\usepackage{multirow}
\usepackage{graphicx}
\usepackage{xcolor,colortbl}
\usepackage{booktabs}
\usepackage{url}
\usepackage{enumitem}
\usepackage{booktabs}
\usepackage{multirow}
\usepackage{graphicx} 
\usepackage[table]{xcolor}
\usepackage{colortbl}

\definecolor{lightblue}{RGB}{235,240,250}

\usepackage[utf8]{inputenc}
\usepackage{afterpage}

\usepackage{microtype}

\usepackage{inconsolata}

\usepackage{booktabs}
\usepackage{multirow}
\usepackage{array}
\usepackage{graphicx}
\usepackage{xcolor,colortbl}

\usepackage{pifont}   
\usepackage{amsfonts}

\title{MM-ShiftKV: Decode-Aware Prefill-Stage KV Selection for Multimodal Large Language Models}

\author{
  \textbf{Jinsong Shu}\textsuperscript{1,}\thanks{~~Equal contribution.}, 
  \textbf{Chenyang Wu}\textsuperscript{1,}\footnotemark[1], 
  \textbf{Zhongle Xie}\textsuperscript{1,\ding{41}}, 
  \textbf{Baokun Wang}\textsuperscript{2}, 
  \textbf{Lidan Shou}\textsuperscript{3,4} \\
  \textsuperscript{1}Zhejiang University \quad
  \textsuperscript{2}Ant Group \\
  \textsuperscript{3}The State Key Laboratory of Blockchain and Data Security, Zhejiang University \\
  \textsuperscript{4}Hangzhou High-Tech Zone (Binjiang) Institute of Blockchain and Data Security
}

\begin{document}
\maketitle

\begin{abstract}


Key-Value (KV) caching is essential for efficient inference in multimodal large language models (MLLMs), yet its memory footprint grows linearly with context length and becomes a major bottleneck due to the large number of visual tokens.
Recent prefill-stage KV selection methods estimate KV importance from prefilling statistics, implicitly assuming that prefilling-time queries are representative of those encountered during decoding.
We show that this assumption breaks down in multimodal inference, where decoding-time queries exhibit substantially larger variance than prefilling-stage representations, leading to unstable KV importance estimation under tight cache budgets.
As a result, small ranking errors can disproportionately discard semantically critical visual tokens and degrade grounding and reasoning performance.
We propose \textbf{MM-ShiftKV}, a training-free, decode-aware and strictly prefill-only KV selection method.
MM-ShiftKV approximates decoding-time query behavior during prefilling by constructing variance-expanded \emph{query proxies} and estimates prompt KV importance based on their aggregated attention mass. Experiments on multimodal benchmarks demonstrate that MM-ShiftKV consistently outperforms existing methods under strict KV-cache budgets. Our code is available at \url{https://github.com/zjuDBxAI/MM-ShiftKV}.

\end{abstract}

\begin{figure*}[t]
  \centering
  \begin{subfigure}[t]{0.61\linewidth}
    \centering
    \includegraphics[width=\linewidth]{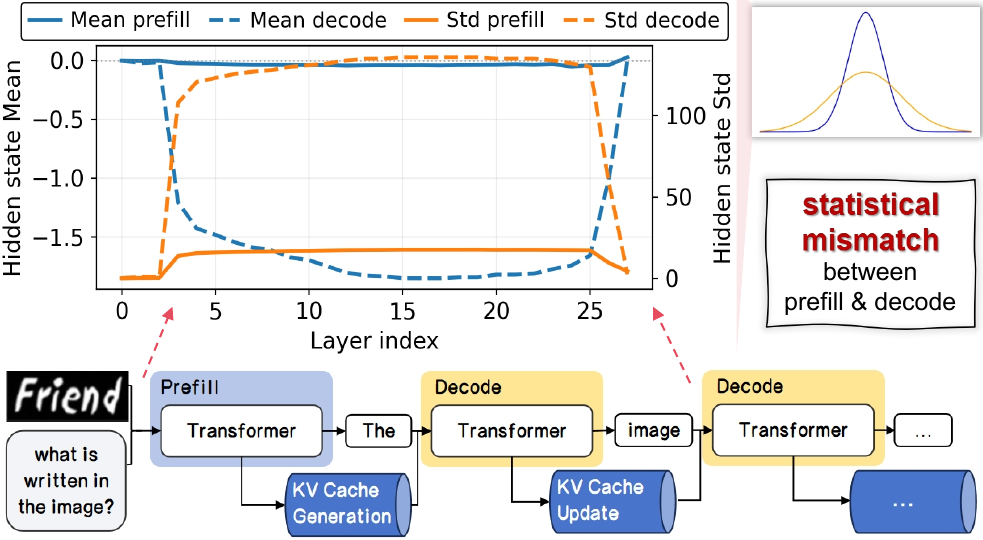}
    \caption{Feature mean/std (OCRBench)}
    \label{fig:mean}
  \end{subfigure}%
  \hspace{0.07\linewidth}
  \begin{subfigure}[t]{0.3\linewidth}
    \centering
    \includegraphics[width=\linewidth]{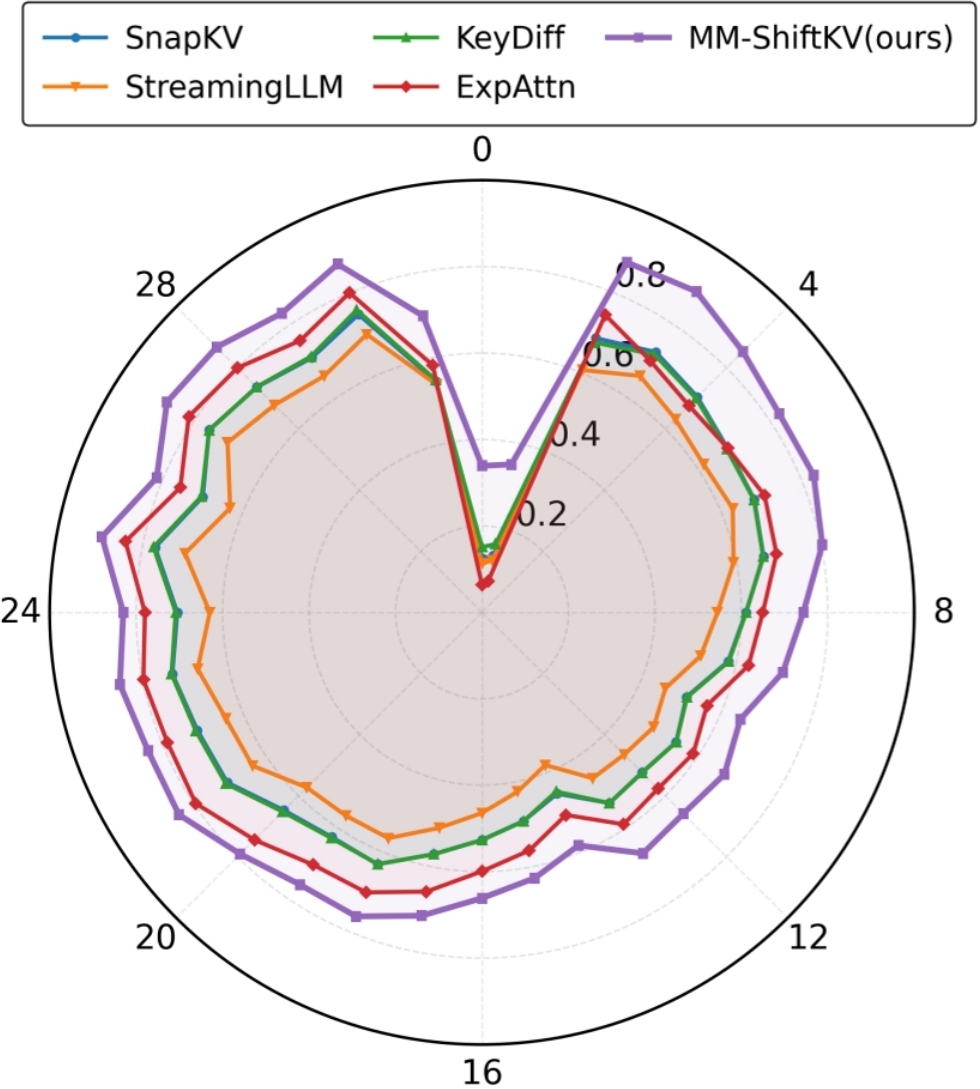}
    \caption{Attention mass coverage}
    \label{fig:overlap}
  \end{subfigure}

\caption{
Prefill--decode statistics and decode-time attention coverage.
\textbf{(a)} Layer-wise mean and variance of hidden-state representations during prefill and decoding on OCRBench.
\textbf{(b)} Attention mass coverage (retained prompt), measured as the fraction of
decode-time attention probability mass assigned to prompt KV tokens retained after prefilling.
}
\end{figure*}

\section{Introduction}
\label{sec:introduction}

Multimodal large language models (MLLMs) extend text-only language models with the ability to generate language grounded in visual inputs, enabling applications such as optical character recognition (OCR), document understanding, and visual question answering~\citep{li2024llavaonevision,bai2025qwen25vl}.
During inference, these models process short textual prompts together with high-resolution visual inputs, which are encoded during prefilling into long multimodal sequences dominated by visual tokens~\citep{arif2025hired}, followed by autoregressive decoding to generate textual outputs.
Efficient decoding relies on \emph{Key--Value (KV) caching}, whose memory footprint grows linearly with the encoded sequence length, making KV cache size and access cost a primary bottleneck for memory consumption and decoding efficiency.

To mitigate this bottleneck, recent work has proposed \emph{prefill-stage KV cache selection}, which retains a subset of KV states after prefilling and reuses them during decoding~\citep{xiao2023efficient,li2024snapkv,devoto2025expected,park2025keydiff}.
Compared to decoding-time cache eviction or adaptive cache compression~\citep{xiao2024infllm}, these prefill-only approaches are attractive because they are training-free~\citep{li2024snapkv}, introduce no decoding-time intervention, and remain compatible with advanced attention kernels such as FlashAttention~\citep{dao2023flashattention2}.
Most methods estimate KV importance from statistics observed during prefilling, implicitly assuming that prefill-stage representations and attention behavior are representative of those encountered during decoding.

This implicit assumption becomes fragile in multimodal inference due to the heterogeneity of multimodal inputs.
A large number of visually redundant tokens coexist with a small subset of semantically critical tokens~\citep{tao2025dycoke,chen2025fastv}, such that small errors in KV ranking may disproportionately remove critical representations.
As a result, existing prefill-stage KV selection methods often lead to degraded language grounding, unstable reasoning, and significant performance drops on multimodal tasks~\citep{devoto2025expected,li2024snapkv,park2025keydiff,devoto2024l2kv}.

At a more fundamental level, prefilling and decoding correspond to distinct functional stages of multimodal inference.
Prefilling primarily emphasizes visual perception and cross-modal alignment, whereas decoding shifts toward language generation and reasoning conditioned on previously generated tokens.
Although the two stages share identical model parameters, they may induce different distributions of hidden states and attention queries, a phenomenon that we systematically analyze in Section~\ref{sec:observation}.
Consequently, importance estimates derived solely from prefilling-stage statistics can be misaligned with decoding-time behavior, leading to unreliable KV selection under strict cache budgets.

In this work, we propose \textbf{MM-ShiftKV}, a training-free and strictly prefill-only KV selection framework for multimodal inference.
Our core idea is to make prefill-stage KV selection explicitly \emph{decode-aware} by calibrating importance estimates to reflect the distributional properties of decoding-time queries, rather than relying solely on prefill-stage statistics.
Decode-aware here does not imply performing KV eviction or re-ranking during decoding, but instead adjusts prefill-based query proxies to better approximate decoding-time behavior.
Concretely, MM-ShiftKV performs one-shot KV selection at the end of prefilling by sampling variance-expanded query proxies from statistics computed over the current input and estimating KV importance based on aggregated attention mass.
The resulting compact prompt KV cache is reused unchanged during decoding, enabling efficient and robust multimodal inference under strict budgets.

\textbf{Contributions.}
\vspace{-0.5em}
\begin{itemize}[leftmargin=1em] 
  \setlength{\itemsep}{0em}
  \setlength{\parskip}{0pt}
  \setlength{\parsep}{0pt}

  \item We identify a persistent prefill--decode \emph{scale mismatch} in
  multimodal inference, where decoding-time \emph{hidden-state representations}
  exhibit substantially larger variance than those observed during prefilling.

  \item We show that this mismatch causes prefill-based query proxies to be
  under-scaled, leading to distorted query--key relevance estimation and
  unstable KV selection under strict KV-cache budgets.

  \item We propose \textbf{MM-ShiftKV}, a training-free and strictly
  \emph{prefill-only} KV selection framework that constructs variance-expanded,
  decode-aware \emph{query proxies} to estimate prompt KV importance.

  \item We demonstrate improved accuracy--memory--latency trade-offs on
  representative OCR, grounding, and long-context VQA benchmarks under tight
  KV-cache constraints.
\end{itemize}

\section{Observation: Prefill--Decode Scale Mismatch in Multimodal Inference}
\label{sec:observation}
Recent work 
has shown that activation statistics in large language models exhibit
structured properties that can be exploited in a training-free manner
\cite{liu2024trainingfree}.
{Inspired by the properties,}
most \emph{prefill-stage} KV selection methods assume that statistics observed during prefilling remain representative of KV usage during decoding.



As discussed in Section~\ref{sec:introduction}, this assumption is critical for prefill-stage KV selection, yet it has not been carefully examined in multimodal inference, and we provide additional theoretical analysis in Appendix \ref{sec:appendix:theory}.
In this section, we show that the assumption is empirically violated and characterize a consistent \emph{prefill-decode statistical mismatch} on \textbf{OCRBench}~\citep{liu2023ocrbench} and \textbf{Qwen2.5-VL-7B-Instruct}. Further results are provided in the Appendix~\ref{sec:appendix:visual}.
\setlength{\parskip}{0pt}
Specifically, we observe that hidden-state representations during prefilling and decoding differ substantially in their statistical \emph{scale}.
Figure~\ref{fig:mean} reports the layer-wise mean and standard deviation of hidden features for \textbf{Qwen2.5-VL-7B-Instruct} evaluated on \textbf{OCRBench}.
Although prefilling and decoding share identical model parameters, decoding-stage representations exhibit consistently larger variance across layers, while mean shifts remain relatively moderate.
\setlength{\parskip}{0pt}
This indicates that prefilling-stage statistics systematically underestimate the scale of decoding-time representations, a phenomenon that lacks dedicated research in prior work on prefill-stage KV selection.

This statistical mismatch has direct implications for the stability of prefill-stage KV selection.
Existing methods typically rely on prefilling-stage signals, such as local attention behavior, sequence-level activation statistics, and KV similarity, to estimate the importance of the token.
When these signals are under-scaled relative to true decoding-time queries, the resulting importance estimates become distorted.
Figure~\ref{fig:overlap} quantifies this effect using \emph{attention mass coverage}, measured as the fraction of decoding-time attention probability mass assigned to prompt KV tokens retained after prefilling.
Following the standard prefill-only protocol, KV selection is performed once at the end of prefilling, and the compressed prompt KV cache is kept fixed during decoding.
Across methods, attention coverage is consistently reduced and exhibits high variance under tight cache budgets, indicating that KV sets selected from prefilling statistics fail to reliably capture decoding-time KV usage.


Taken together, these results demonstrate that prefill-stage KV selection based solely on prefilling-stage statistics is inherently brittle in multimodal settings.
The observed prefill-decode statistical mismatch highlights the need to explicitly account for decoding-time query behavior while remaining strictly within the prefill-only regime.
This insight forms the basis for the decode-aware prefill-stage KV selection approach developed in the next section, which achieves the highest attention coverage shown in Figure~\ref{fig:overlap}.

\begin{figure*}[t]
  \centering
  \includegraphics[width=0.8\textwidth]{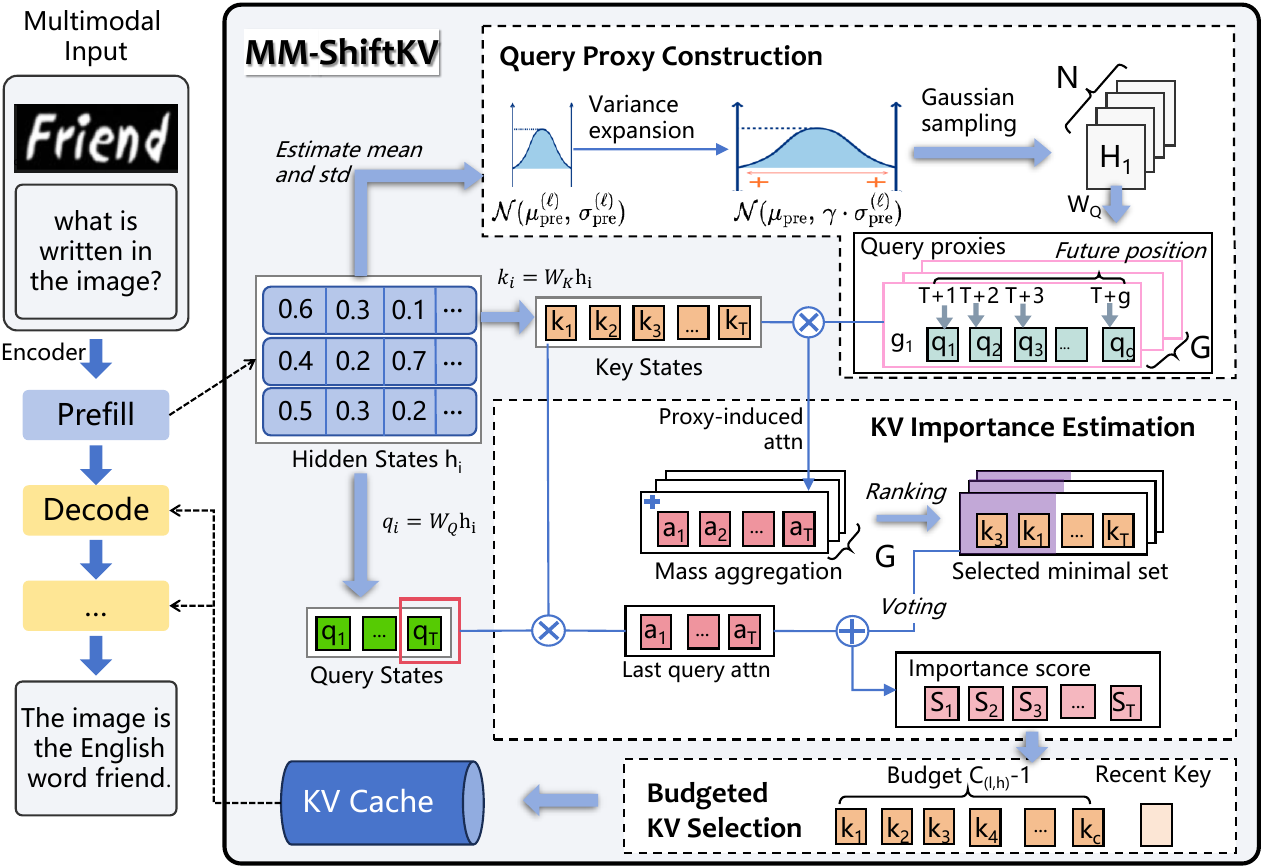}
\caption{
Overview of MM-ShiftKV.
The method computes statistics over the full prefill sequence of the current input
and constructs variance-expanded, decode-aware \emph{query proxies} during prefilling.
These query proxies are used to estimate prompt KV importance via attention-mass
aggregation and group-wise voting with a last-token anchor, yielding a compact
KV cache under a fixed budget.
}

  \label{fig:mmshiftkv}
\end{figure*}

\section{Method}
\label{sec:method}

\paragraph{Core Idea.}
MM-ShiftKV addresses the instability of prefill-stage KV selection in multimodal inference by explicitly approximating how prompt keys will be accessed by \emph{future decoding queries}.
Since decoding-time queries are unavailable during prefilling, the core idea is to construct a set of synthetic \emph{query proxies} during the prefill stage that approximate the distributional properties of decoding queries.
Prompt keys that are consistently attended by these query proxies are more likely to be important during decoding and should therefore be retained under the limited KV-cache budget.

\subsection{Problem Definition}
\label{sec:method:formulation}
Before describing our method, we formalize the prefill-stage KV selection problem in multimodal inference.

\paragraph{Input Sequence.}
Given a multimodal input sequence
\begin{equation}
X = (x_1, x_2, \ldots, x_T),
\end{equation}
where tokens include both visual and textual modalities, and $T$ denotes the sequence length of the \emph{prefill stage}.

\paragraph{Prompt KV cache.}
During prefilling, the model computes KV representations for all prompt tokens at each transformer layer $\ell$ and KV head $h$.
The resulting set of KV pairs forms the \emph{prompt KV cache}:
\begin{equation}
\mathcal{C}^{(\ell,h)} =
\{(k_t^{(\ell,h)}, v_t^{(\ell,h)})\}_{t \in \mathcal{T}},
\end{equation}
where $\mathcal{T} = \{1,\ldots,T\}$ indexes prompt tokens.
These prompt KVs are repeatedly accessed during decoding and dominate memory consumption and attention computation in inference.

\paragraph{Prefill-stage KV selection.}
With a given cache budget $C_{\ell,h}$ for each layer $\ell$ and KV head $h$, the objective of the prefill-stage KV selection is to retain a subset
\begin{equation}
\mathcal{C'}^{(\ell,h)} \subseteq \mathcal{C}^{(\ell,h)}, \qquad
|\mathcal{C'}^{(\ell,h)}| \le C_{\ell,h},
\end{equation}
which is reused throughout the decoding to reduce memory usage and attention cost while preserving generation quality.

In practice, the KV selection is performed independently for each layer and head, and executed once at the end of \emph{prefilling}.
Note that the selected prompt KV cache remains fixed during decoding, while KV pairs from newly generated tokens are appended in a standard autoregressive manner.

\subsection{MM-ShiftKV}
\label{sec:method:overview}

Figure~\ref{fig:mmshiftkv} provides an overview of MM-ShiftKV.
The method consists of three steps executed during prefilling:
(1) constructing decode-aware query proxies,
(2) estimating prompt KV importance via mass-based voting,
and (3) performing budgeted KV selection based on aggregated importance scores.
All steps are strictly prefill-only and introduce no decoding-time intervention. More implementation details can be found in Appendix \ref{sec:appendix:implementation}.

\subsubsection{Query Proxy Construction}
\label{sec:method:proxy}

To approximate decoding-time query behavior, MM-ShiftKV constructs a set of synthetic \emph{query proxies} using prefilling-stage statistics.

\paragraph{Prefilling hidden-state statistics.}
Let $h^{(\ell)} \in \mathbb{R}^d$ denote the hidden states produced at layer $\ell$ during prefilling.
We summarize these representations using element-wise statistics:
\begin{equation}
\mu_{\text{pre}}^{(\ell)} = \mathrm{mean}(h^{(\ell)}), \qquad
\sigma_{\text{pre}}^{(\ell)} = \mathrm{std}(h^{(\ell)}),
\end{equation}
computed across prompt tokens for each feature dimension. These statistics are computed separately for each input sample over its full prefill sequence, rather than shared across different samples.

\paragraph{Variance-expanded sampling.}
As shown in Section~\ref{sec:observation}, decoding-time queries exhibit substantially larger variance than prefilling-stage representations.
To approximate this effect, we introduce a variance expansion factor $\gamma > 1$ and sample $N$ \emph{proxy hidden states} $\{\tilde{H}_i^{(\ell)}\}_{i=1}^{N}$ from
\begin{equation}
\tilde{H}_i^{(\ell)} \sim
\mathcal{N}\!\left(
\mu_{\text{pre}}^{(\ell)},
\ \mathrm{diag}\!\left((\gamma \sigma_{\text{pre}}^{(\ell)})^2\right)
\right).
\end{equation}

\paragraph{Query proxy projection and positioning.}
Each proxy hidden state is projected into the query space using the model’s query projection matrix:
\begin{equation}
\tilde{q}_i^{(\ell,h)} = W_Q^{(\ell,h)} \tilde{H}_i^{(\ell)}, 
\qquad i = 1,\ldots,N.
\end{equation}
The resulting vectors $\{\tilde{q}_i^{(\ell,h)}\}$ are referred to as \emph{query proxies}.
To reflect that decoding queries attend to prompt keys from future positions, query proxies are assigned synthetic future positions and encoded using the model’s rotary positional embedding (RoPE), while prompt keys retain their prefilling-stage positional encodings.

\subsubsection{KV Importance Estimation}
\label{sec:method:importance}

Given a set of query proxies, MM-ShiftKV estimates the importance of each prompt key based on how consistently it attends across diverse proxies.

\paragraph{Proxy-induced attention.}
For each query proxy $\tilde{q}_i^{(\ell,h)}$, we compute its attention distribution over prompt keys:
\begin{equation}
a_t\!\left(\tilde{q}_i^{(\ell,h)}\right) =
\mathrm{softmax}\!\left(
\frac{\left(\tilde{q}_i^{(\ell,h)}\right)^\top k_t^{(\ell,h)}}{\sqrt{d}}
\right),
t \in \mathcal{T}
\end{equation}

\paragraph{Mass aggregation and voting.}
The $N$ query proxies are partitioned into $G$ disjoint groups $\{\mathcal{I}_{g'}\}_{g'=1}^{G}$.
For each group $g'$, attention masses are aggregated as
\begin{equation}
\bar{a}_t^{(g')} =
\sum_{i \in \mathcal{I}_{g'}} 
a_t\!\left(\tilde{q}_i^{(\ell,h)}\right),
\quad t \in \mathcal{T}
\end{equation}
Prompt keys are then ranked in descending order of $\bar{a}_t^{(g')}$, and we define $\mathcal{S}_{g'} \subseteq \mathcal{T}$ as the \emph{smallest} set of keys whose cumulative mass satisfies
\begin{equation}
\sum_{t \in \mathcal{S}_{g'}} \bar{a}_t^{(g')}
\;\ge\;
\tau \sum_{t \in \mathcal{T}} \bar{a}_t^{(g')},
\end{equation}
where $\tau \in (0,1)$ is a predefined mass threshold.
Each selected key receives one vote, and the final importance score of key $t$ is given by
\begin{equation}
\mathrm{vote}(t) = \sum_{g'=1}^{G} \mathbf{1}[t \in \mathcal{S}_{g'}].
\end{equation}

\subsubsection{Budgeted KV Selection}
\label{sec:method:selection}

KV selection is performed independently for each layer $\ell$ and KV head $h$.
Among the prompt keys, we retain the most recent key to ensure decoding stability and select the remaining keys based on their aggregated importance scores.
To make ranking deterministic when vote counts tie, we use a last-query anchor:
\begin{equation}
S_t^{(\ell,h)}=\mathrm{vote}(t)+\lambda\cdot a_t\!\left(q_{\mathrm{last}}^{(\ell,h)}\right),
\end{equation}
where $a_t(q_{\mathrm{last}}^{(\ell,h)})$ is the attention score from the last
real prefill query to key $k_t^{(\ell,h)}$, and $\lambda$ is a small constant
($\lambda{=}1$ by default).
The ranking follows a lexicographical-like priority:
the discrete vote count is primary, while the anchor term is secondary.
Because the anchor weight does not exceed one integer vote step, it only acts as
a tie-breaker and does not override proxy voting.
Specifically, the top $C_{\ell,h}-1$ keys ranked by $S_t^{(\ell,h)}$ are used to
form the compressed prompt KV cache, and the most recent token is always kept.

\paragraph{Overhead.} Similar to existing prefill-only approaches, MM-ShiftKV introduces additional computation only during the \emph{prefill stage}.
For each layer and KV head, it computes attention between $N$ query proxies and $|\mathcal{T}|$ prompt keys once, followed by group-wise aggregation.
The time complexity scales as $O(N|\mathcal{T}|)$, while the peak additional memory can be bounded by $O(|\mathcal{T}|)$ using streaming implementations.

\begin{table*}[t!]
\centering
\scriptsize
\setlength{\tabcolsep}{3pt}
\renewcommand{\arraystretch}{1.05}
\huge
\resizebox{\textwidth}{!}{%
\begin{tabular}{
l|
c >{\columncolor{lightblue}}c c >{\columncolor{lightblue}}c|
c >{\columncolor{lightblue}}c c >{\columncolor{lightblue}}c|
c >{\columncolor{lightblue}}c c >{\columncolor{lightblue}}c|
c >{\columncolor{lightblue}}c c >{\columncolor{lightblue}}c|
c >{\columncolor{lightblue}}c c >{\columncolor{lightblue}}c|
c >{\columncolor{lightblue}}c c >{\columncolor{lightblue}}c|
c >{\columncolor{lightblue}}c c >{\columncolor{lightblue}}c
}

\toprule
\multirow{2}{*}{\textbf{Method}}
& \multicolumn{4}{c|}{\textbf{DocVQA}}
& \multicolumn{4}{c|}{\textbf{OCRBench}}
& \multicolumn{4}{c|}{\textbf{TextVQA}}
& \multicolumn{4}{c|}{\textbf{ChartQA}}
& \multicolumn{4}{c|}{\textbf{TextCaps}}
& \multicolumn{4}{c|}{\textbf{MMMU}}
& \multicolumn{4}{c}{\textbf{Avg}} \\

\cline{2-29}

& \textbf{64} & \textbf{128} & \textbf{256} & \textbf{512}
& \textbf{64} & \textbf{128} & \textbf{256} & \textbf{512}
& \textbf{64} & \textbf{128} & \textbf{256} & \textbf{512}
& \textbf{64} & \textbf{128} & \textbf{256} & \textbf{512}
& \textbf{64} & \textbf{128} & \textbf{256} & \textbf{512}
& \textbf{64} & \textbf{128} & \textbf{256} & \textbf{512}
& \textbf{64} & \textbf{128} & \textbf{256} & \textbf{512} \\
\midrule

FullKV
& \multicolumn{4}{c|}{94.5}
& \multicolumn{4}{c|}{82.3}
& \multicolumn{4}{c|}{83.0}
& \multicolumn{4}{c|}{83.2}
& \multicolumn{4}{c|}{58.7}
& \multicolumn{4}{c|}{50.8}
& \multicolumn{4}{c}{75.4} \\

StreamingLLM
& 41.0 & 45.1 & 56.5 & 69.1
& 29.3 & 48.5 & 65.2 & 72.0
& 52.4 & 60.6 & 69.4 & 76.8
& 68.7 & 74.4 & 80.2 & 82.3
& 22.1 & 34.3 & 45.2 & 53.2
& 50.2 & 50.4 & 50.2 & 50.4
& 44.0 & 52.2 & 61.1 & 67.3 \\

ExpectedAttn
& 56.3 & 59.7 & 66.2 & 77.6
& 57.5 & 67.3 & 72.2 & 77.9
& 64.2 & 69.4 & 75.6 & 81.3
& 77.5 & 77.9 & 80.5 & 83.0
& 40.1 & 48.1 & 53.7 & 57.6
& 50.9 & 51.0 & 50.9 & 50.9
& 57.8 & 62.2 & 66.5 & 71.4 \\

SnapKV
& 75.6 & 88.8 & 93.0 & 94.1
& 53.5 & 72.1 & 79.1 & 81.1
& 69.8 & 78.7 & 81.6 & 82.9
& 80.0 & 84.1 & 83.6 & 83.2
& 29.3 & 46.6 & 55.9 & 58.9
& 50.7 & 50.2 & 50.6 & 50.4
& 59.8 & 70.1 & 74.0 & 75.1 \\

KEYDIFF
& 51.0 & 61.5 & 74.7 & 86.7
& 41.7 & 62.7 & 74.8 & 79.1
& 72.1 & 78.6 & 82.0 & 83.0
& 73.0 & 80.8 & 84.1 & 83.2
& 42.2 & 49.9 & 55.2 & 59.2
& 50.9 & 50.7 & 50.6 & 50.4
& 55.2 & 64.0 & 70.2 & 73.6 \\

\textbf{MM-ShiftKV}
& \textbf{81.6} & \textbf{88.9} & \textbf{92.8} & \textbf{94.3}
& \textbf{68.8} & \textbf{76.1} & \textbf{80.2} & \textbf{81.9}
& \textbf{80.0} & \textbf{82.2} & \textbf{82.8} & \textbf{82.9}
& \textbf{84.4} & \textbf{84.3} & \textbf{83.7} & \textbf{83.3}
& \textbf{50.4} & \textbf{58.9} & \textbf{60.0} & \textbf{60.1}
& \textbf{50.6} & \textbf{50.4} & \textbf{50.4} & \textbf{50.4}
& \textbf{69.3} & \textbf{73.5} & \textbf{75.0} & \textbf{75.5} \\

\bottomrule
\end{tabular}%
}

\caption{
Results on multimodal benchmarks under different per-head KV-cache budgets
(64/128/256/512) using \textbf{Qwen2.5-VL-7B-Instruct}.
FullKV denotes standard inference without KV cache compression.
Avg is the arithmetic mean over DocVQA, OCRBench, TextVQA, ChartQA, TextCaps, and MMMU.
}

\label{tab:mm_budget_qwen}
\end{table*}

\begin{table*}[t!]
\centering
\scriptsize
\setlength{\tabcolsep}{3pt}
\renewcommand{\arraystretch}{1.05}
\huge
\resizebox{\textwidth}{!}{%
\begin{tabular}{
l|
c >{\columncolor{lightblue}}c c >{\columncolor{lightblue}}c|
c >{\columncolor{lightblue}}c c >{\columncolor{lightblue}}c|
c >{\columncolor{lightblue}}c c >{\columncolor{lightblue}}c|
c >{\columncolor{lightblue}}c c >{\columncolor{lightblue}}c|
c >{\columncolor{lightblue}}c c >{\columncolor{lightblue}}c|
c >{\columncolor{lightblue}}c c >{\columncolor{lightblue}}c|
c >{\columncolor{lightblue}}c c >{\columncolor{lightblue}}c
}

\toprule
\multirow{2}{*}{\textbf{Method}}
& \multicolumn{4}{c|}{\textbf{DocVQA}}
& \multicolumn{4}{c|}{\textbf{OCRBench}}
& \multicolumn{4}{c|}{\textbf{TextVQA}}
& \multicolumn{4}{c|}{\textbf{ChartQA}}
& \multicolumn{4}{c|}{\textbf{TextCaps}}
& \multicolumn{4}{c|}{\textbf{MMMU}}
& \multicolumn{4}{c}{\textbf{Avg}} \\

\cline{2-29}

& \textbf{64} & \textbf{128} & \textbf{256} & \textbf{512}
& \textbf{64} & \textbf{128} & \textbf{256} & \textbf{512}
& \textbf{64} & \textbf{128} & \textbf{256} & \textbf{512}
& \textbf{64} & \textbf{128} & \textbf{256} & \textbf{512}
& \textbf{64} & \textbf{128} & \textbf{256} & \textbf{512}
& \textbf{64} & \textbf{128} & \textbf{256} & \textbf{512}
& \textbf{64} & \textbf{128} & \textbf{256} & \textbf{512} \\
\midrule

FullKV
& \multicolumn{4}{c|}{68.0}
& \multicolumn{4}{c|}{52.0}
& \multicolumn{4}{c|}{65.0}
& \multicolumn{4}{c|}{55.0}
& \multicolumn{4}{c|}{73.0}
& \multicolumn{4}{c|}{36.4}
& \multicolumn{4}{c}{58.2} \\

StreamingLLM
& 27.3 & 28.6 & 30.8 & 35.8
& 12.9 & 19.4 & 25.4 & 31.8
& 39.4 & 41.0 & 42.8 & 47.1
& 23.9 & 23.9 & 24.3 & 27.1
& 28.8 & 33.0 & 36.2 & 42.1
& 36.9 & 36.4 & 36.4 & 36.2
& 28.2 & 30.4 & 32.7 & 36.7 \\

ExpectedAttn
& 40.6 & 47.0 & 54.8 & 62.1
& 32.3 & 37.9 & 43.5 & 47.9
& 54.2 & 57.2 & 60.5 & 62.6
& 39.0 & 43.8 & 49.7 & 51.8
& 54.9 & 60.0 & 64.6 & 67.0
& 36.1 & 36.6 & 36.6 & 36.6
& 42.9 & 47.1 & 51.6 & 54.7 \\

SnapKV
& 49.7 & 59.0 & 64.3 & 67.0
& 33.5 & 40.8 & 46.2 & 51.1
& 56.1 & 60.1 & 62.2 & 64.1
& 42.9 & 45.8 & 49.8 & 54.2
& 44.2 & 56.1 & 65.3 & 69.7
& 36.4 & 36.7 & 36.4 & 36.7
& 43.8 & 49.8 & 54.0 & 57.1 \\

KEYDIFF
& 43.0 & 49.2 & 56.5 & 62.6
& 29.8 & 36.8 & 43.5 & 48.4
& 57.7 & 60.3 & 63.1 & 64.0
& 40.1 & 44.5 & 49.6 & 52.3
& 59.7 & 66.3 & 70.7 & 73.8
& 36.3 & 36.6 & 36.7 & 36.8
& 44.4 & 49.0 & 53.4 & 56.3 \\

\textbf{MM-ShiftKV}
& \textbf{56.3} & \textbf{61.8} & \textbf{65.4} & \textbf{67.1}
& \textbf{41.1} & \textbf{45.7} & \textbf{49.4} & \textbf{51.9}
& \textbf{62.1} & \textbf{62.8} & \textbf{64.2} & \textbf{64.6}
& \textbf{51.5} & \textbf{52.1} & \textbf{52.6} & \textbf{54.5}
& \textbf{63.2} & \textbf{68.9} & \textbf{73.1} & \textbf{74.8}
& \textbf{36.6} & \textbf{36.4} & \textbf{36.6} & \textbf{36.6}
& \textbf{51.8} & \textbf{54.6} & \textbf{56.9} & \textbf{58.3} \\

\bottomrule
\end{tabular}%
}

\caption{
Results on multimodal benchmarks under different per-head KV-cache budgets
(64/128/256/512) using \textbf{LLaVA-v1.6-Vicuna-7B}.
FullKV denotes standard inference without KV cache compression.
Avg is the arithmetic mean over DocVQA, OCRBench, TextVQA, ChartQA, TextCaps, and MMMU.
}
\label{tab:mm_budget_llava}
\end{table*}

\begin{table*}[t!]
\centering
\scriptsize
\setlength{\tabcolsep}{3pt}
\renewcommand{\arraystretch}{1.05}
\huge
\resizebox{\textwidth}{!}{%
\begin{tabular}{
l|
c >{\columncolor{lightblue}}c c >{\columncolor{lightblue}}c|
c >{\columncolor{lightblue}}c c >{\columncolor{lightblue}}c|
c >{\columncolor{lightblue}}c c >{\columncolor{lightblue}}c|
c >{\columncolor{lightblue}}c c >{\columncolor{lightblue}}c|
c >{\columncolor{lightblue}}c c >{\columncolor{lightblue}}c|
c >{\columncolor{lightblue}}c c >{\columncolor{lightblue}}c|
c >{\columncolor{lightblue}}c c >{\columncolor{lightblue}}c
}

\toprule
\multirow{2}{*}{\textbf{Method}}
& \multicolumn{4}{c|}{\textbf{DocVQA}}
& \multicolumn{4}{c|}{\textbf{OCRBench}}
& \multicolumn{4}{c|}{\textbf{TextVQA}}
& \multicolumn{4}{c|}{\textbf{ChartQA}}
& \multicolumn{4}{c|}{\textbf{TextCaps}}
& \multicolumn{4}{c|}{\textbf{MMMU}}
& \multicolumn{4}{c}{\textbf{Avg}} \\

\cline{2-29}

& \textbf{64} & \textbf{128} & \textbf{256} & \textbf{512}
& \textbf{64} & \textbf{128} & \textbf{256} & \textbf{512}
& \textbf{64} & \textbf{128} & \textbf{256} & \textbf{512}
& \textbf{64} & \textbf{128} & \textbf{256} & \textbf{512}
& \textbf{64} & \textbf{128} & \textbf{256} & \textbf{512}
& \textbf{64} & \textbf{128} & \textbf{256} & \textbf{512}
& \textbf{64} & \textbf{128} & \textbf{256} & \textbf{512} \\

\midrule

PyramidKV
& 49.0 & 60.0 & 64.9 & 67.1
& 31.2 & 40.9 & 47.6 & 50.1
& 55.0 & 60.0 & 62.6 & 64.1
& 41.0 & 47.1 & 53.0 & 54.9
& 41.6 & 57.4 & 64.9 & 69.5
& 36.3 & 36.4 & 36.4 & 36.4
& 42.3 & 50.3 & 54.9 & 57.0 \\

\qquad \textbf{+ours}
& \textbf{\textcolor{red}{$\uparrow$}\,8.9}  & \textbf{\textcolor{red}{$\uparrow$}\,2.5}  & \textbf{\textcolor{red}{$\uparrow$}\,1.0}  & \textbf{\textcolor{red}{$\uparrow$}\,0.1}
& \textbf{\textcolor{red}{$\uparrow$}\,11.6} & \textbf{\textcolor{red}{$\uparrow$}\,6.5}  & \textbf{\textcolor{red}{$\uparrow$}\,1.9}  & \textbf{\textcolor{red}{$\uparrow$}\,1.2}
& \textbf{\textcolor{red}{$\uparrow$}\,6.4}  & \textbf{\textcolor{red}{$\uparrow$}\,3.3}  & \textbf{\textcolor{red}{$\uparrow$}\,1.4}  & \textbf{\textcolor{red}{$\uparrow$}\,0.5}
& \textbf{\textcolor{red}{$\uparrow$}\,8.4}  & \textbf{\textcolor{red}{$\uparrow$}\,5.6}  & \textbf{\textcolor{red}{$\uparrow$}\,1.7}  & \textbf{\textcolor{red}{$\uparrow$}\,0.1}
& \textbf{\textcolor{red}{$\uparrow$}\,23.2} & \textbf{\textcolor{red}{$\uparrow$}\,11.5} & \textbf{\textcolor{red}{$\uparrow$}\,8.2}  & \textbf{\textcolor{red}{$\uparrow$}\,4.8}
& \textbf{\textcolor{red}{$\uparrow$}\,0.1} & \textbf{0.0} & \textbf{\textcolor{red}{$\uparrow$}\,0.1}  & \textbf{\textcolor{red}{$\uparrow$}\,0.1}
& \textbf{\textcolor{red}{$\uparrow$}\,9.8} & \textbf{\textcolor{red}{$\uparrow$}\,4.9} & \textbf{\textcolor{red}{$\uparrow$}\,2.4}  & \textbf{\textcolor{red}{$\uparrow$}\,1.1} \\

AdaKV
& 54.1 & 60.6 & 65.2 & 67.5
& 35.1 & 43.0 & 48.4 & 50.8
& 56.0 & 59.8 & 61.9 & 64.2
& 43.8 & 45.7 & 49.1 & 54.8
& 47.2 & 59.2 & 65.9 & 70.3
& 36.1 & 36.3 & 36.4 & 36.4
& 45.4 & 50.8 & 54.5 & 57.3 \\

\qquad \textbf{+ours}
& \textbf{\textcolor{red}{$\uparrow$}\,4.7}  & \textbf{\textcolor{red}{$\uparrow$}\,3.4}  & \textbf{\textcolor{red}{$\uparrow$}\,0.7}  & \textbf{0.0}
& \textbf{\textcolor{red}{$\uparrow$}\,9.2}  & \textbf{\textcolor{red}{$\uparrow$}\,4.1}  & \textbf{\textcolor{red}{$\uparrow$}\,2.6}  & \textbf{\textcolor{red}{$\uparrow$}\,1.9}
& \textbf{\textcolor{red}{$\uparrow$}\,5.1}  & \textbf{\textcolor{red}{$\uparrow$}\,2.0}  & \textbf{\textcolor{red}{$\uparrow$}\,1.5}  & \textbf{\textcolor{red}{$\uparrow$}\,0.5}
& \textbf{\textcolor{red}{$\uparrow$}\,5.1}  & \textbf{\textcolor{red}{$\uparrow$}\,3.8}  & \textbf{\textcolor{red}{$\uparrow$}\,3.4}  & \textbf{0.0}
& \textbf{\textcolor{red}{$\uparrow$}\,19.0} & \textbf{\textcolor{red}{$\uparrow$}\,11.7} & \textbf{\textcolor{red}{$\uparrow$}\,7.2}  & \textbf{\textcolor{red}{$\uparrow$}\,4.3}
& \textbf{\textcolor{red}{$\uparrow$}\,0.2} & \textbf{\textcolor{red}{$\uparrow$}\,0.1} & \textbf{0.0}  & \textbf{0.0}
& \textbf{\textcolor{red}{$\uparrow$}\,7.2}  & \textbf{\textcolor{red}{$\uparrow$}\,4.2}  & \textbf{\textcolor{red}{$\uparrow$}\,2.6}  & \textbf{\textcolor{red}{$\uparrow$}\,1.1} \\

SparseMM
& 63.7 & 66.7 & 67.8 & 68.0
& 47.8 & 50.1 & 52.3 & 52.8
& 63.1 & 64.1 & 64.7 & 64.7
& 52.7 & 53.8 & 53.9 & 54.8
& 59.9 & 70.7 & 72.8 & 73.3
& 36.4 & 36.6 & 36.3 & 36.4
& 53.9 & 57.0 & 58.0 & 58.3 \\

\qquad \textbf{+ours}
& \textbf{\textcolor{red}{$\uparrow$}\,1.2} & \textbf{\textcolor{red}{$\uparrow$}\,0.3} & \textbf{\textcolor{red}{$\uparrow$}\,0.1} & \textbf{0.0}
& \textbf{\textcolor{red}{$\uparrow$}\,1.3} & \textbf{\textcolor{red}{$\uparrow$}\,0.7} & \textbf{\textcolor{red}{$\uparrow$}\,0.1} & \textbf{0.0}
& \textbf{\textcolor{red}{$\uparrow$}\,0.4} & \textbf{0.0} & \textbf{0.0} & \textbf{0.0}
& \textbf{\textcolor{red}{$\uparrow$}\,0.1} & \textbf{\textcolor{red}{$\uparrow$}\,0.1} & \textbf{\textcolor{red}{$\uparrow$}\,0.5} & \textbf{0.0}
& \textbf{\textcolor{red}{$\uparrow$}\,6.0} & \textbf{0.0} & \textbf{\textcolor{red}{$\uparrow$}\,0.1} & \textbf{\textcolor{red}{$\uparrow$}\,0.1}
& \textbf{0.0} & \textbf{\textcolor{teal}{$\downarrow$}\,0.2} & \textbf{\textcolor{red}{$\uparrow$}\,0.1}  & \textbf{0.0}
& \textbf{\textcolor{red}{$\uparrow$}\,1.5} & \textbf{\textcolor{red}{$\uparrow$}\,0.2} & \textbf{\textcolor{red}{$\uparrow$}\,0.2} & \textbf{0.0} \\

\bottomrule
\end{tabular}%
}

\caption{
Results on multimodal benchmarks with different KV-cache \emph{budget allocation}
methods under various per-head KV-cache budgets (64/128/256/512) using
\textbf{LLaVA-v1.6-Vicuna-7B}.
\textbf{+ours} applies MM-ShiftKV as a prefill-only KV selection module on top of
the original allocation strategy, without changing its budget.
Avg is the arithmetic mean over DocVQA, OCRBench, TextVQA, ChartQA, TextCaps, and MMMU.
}
\label{tab:mm_budget_allocation_llava}
\end{table*}

\section{Experiments}
\label{sec:experiments}

We conduct extensive experiments to evaluate \textbf{MM-ShiftKV} under
memory-constrained multimodal inference.
Following prior work on prompt KV cache compression
(e.g., SnapKV~\citep{li2024snapkv}, ExpectedAttn~\citep{devoto2025expected}, and KEYDIFF~\citep{park2025keydiff}),
we focus on a \emph{one-shot prefilling} setting,
where the KV cache is compressed once after prompt encoding
and kept fixed throughout decoding.

Our experiments aim to answer the following questions:
(i) how MM-ShiftKV compares with existing \emph{prefill-stage} KV compression
baselines under strict KV-cache budgets,
(ii) how performance degrades as the cache budget decreases, and
(iii) what accuracy-memory-latency trade-offs can be achieved while remaining
compatible with FlashAttention~\citep{dao2023flashattention2}-style decoding kernels.

\subsection{Experimental Setup}

\label{sec:exp:setup}
\paragraph{Models, Datasets, and Metrics.}
We evaluate MM-ShiftKV on two representative multimodal large language models, \textbf{Qwen2.5-VL-7B-Instruct}~\citep{bai2025qwen25vl} and \textbf{LLaVA-v1.6-Vicuna-7B}~\citep{li2024llavaonevision}.
Experiments are conducted on a diverse suite of multimodal benchmarks covering document understanding, OCR-centric visual question answering, chart reasoning, and image captioning.
We employ \textbf{OCRBench}~\citep{liu2023ocrbench} using exact-match accuracy to measure precise text recognition and \textbf{DocVQA}~\citep{mathew2021docvqa} using Average Normalized Levenshtein Similarity (ANLS) as the metric.
For chart reasoning and visual question answering, we use exact-match accuracy on \textbf{ChartQA}~\citep{masry2022chartqa}, \textbf{TextVQA}~\citep{singh2019towards}, and \textbf{MMMU}~\citep{yue2024mmmu} to measure the fraction of correctly answered questions.
For image captioning, we evaluate \textbf{TextCaps}~\citep{sidorov2020textcaps} using caption quality consensus with reference captions (CIDEr) as the metric.
The average number of input tokens for each benchmark is reported in Table~\ref{tab:token_stats}. Consistent with the observations in \citet{wang2025sparsemm}, we find that while text instructions remain concise, the inclusion of high-resolution visual tokens significantly expands the sequence length, posing a primary bottleneck for KV cache memory.
For all metrics, higher values indicate better performance.

\paragraph{Baselines.}
We compare against eviction-free inference (\textbf{Full KV}) and strong
prefilling-stage KV compression baselines, including \textbf{SnapKV}~\citep{li2024snapkv},
\textbf{ExpectedAttn}~\citep{devoto2025expected}, \textbf{StreamingLLM}~\citep{xiao2023efficient}, and \textbf{KEYDIFF}~\citep{park2025keydiff} (one-shot variant).
All baselines are evaluated under identical KV-cache budgets and decoding
settings, using their recommended configurations.
More details are provided in the Appendix~\ref{sec:appendix:datails}.

\paragraph{Prefill-Only Evaluation Protocol and KV Budget.}
Following prior work 
, we adopt a one-shot
prefilling protocol.
Given an input prompt consisting of both visual and textual tokens, we first
perform standard prefilling to compute the full KV cache.
Each method is then applied once at the end of prefilling to score and select a
subset of KV pairs according to its selection policy, producing a compressed
prompt KV cache that is kept fixed throughout decoding.
No decode-time eviction, re-ranking, or decoding-time statistics are used.

KV-cache budgets are defined at the level of individual KV heads for each transformer layer, with per-head budgets $C\in\{64, 128, 256, 512\}$.
Unless otherwise specified, the same budget is applied uniformly across all layers and all KV heads.
For models with Grouped-Query Attention (GQA)~\citep{ainslie2023gqa} or Multi-Query Attention (MQA)~\citep{komatsuzaki2022sparse},
where multiple query heads share one KV head, attention statistics from the corresponding query heads are aggregated, and the budget is applied at the head level.

\paragraph{Implementation and Hyperparameters.}
Unless otherwise specified, we use $N{=}512$ decode-aware query proxies,
partitioned into $G{=}32$ groups of size $g{=}16$ ($N=Gg$), with variance expansion
factor $\gamma{=}10$, attention mass threshold $\tau{=}0.95$, and last-token
anchor weight $\lambda{=}1$.
All hyperparameters are fixed globally and shared across models, datasets, and
budgets.
We apply greedy decoding with a maximum generation length of 64.
All experiments are conducted on NVIDIA H100 80GB GPUs using CUDA 12.8 and FlashAttention 2.4.1 with mixed-precision (fp16/bf16). Additional empirical analysis are detailed in the Appendix~\ref{sec:appendix:sensitivity}. All results are averaged over 3 independent runs with different random seeds.

\begin{table}[t]
\centering
\small
\setlength{\tabcolsep}{4pt}
\resizebox{\columnwidth}{!}{%
\begin{tabular}{lccccc}
\toprule
Task & OCRBench & DocVQA & ChartQA & TextVQA & TextCaps \\
\midrule
LLaVA-Series & 1700 & 2433 & 2270 & 2376 & 2376 \\
Qwen2.5-VL-7B-Instruct   & 1245 & 4830 & 642  & 1024 & 1024 \\
\bottomrule
\end{tabular}}
\caption{
Average number of input tokens across benchmarks.
Text instructions are short, and visual tokens constitute the majority of the
input sequence.
}
\label{tab:token_stats}
\end{table}

\paragraph{Results on multimodal benchmarks.}
We evaluate \textbf{MM-ShiftKV} on a diverse set of multimodal benchmarks under different per-head KV-cache budgets. 
As shown in Table~\ref{tab:mm_budget_qwen} and Table~\ref{tab:mm_budget_llava}, MM-ShiftKV consistently outperforms existing prefill-stage KV selection baselines across both backbone models (Qwen2.5-VL and LLaVA-v1.6) and all tasks, with the performance gap becoming more pronounced as the KV budget decreases.

Under extreme compression (64 tokens per KV head), baseline methods often suffer from performance degradation, particularly on OCR- and grounding-centric tasks. Compared to strong prefill-only baselines, MM-ShiftKV achieves relative accuracy improvements on the order of \(20\%\)–\(30\%\) in representative document understanding benchmarks. When compared to decode-agnostic or heuristic KV selection methods, relative performance gains can exceed \(50\%\), highlighting the brittleness of decode-unaware KV management under multimodal inputs.

Consistent trends are also observed on generation-oriented tasks. Under the same extreme budget, MM-ShiftKV yields relative improvements of approximately \(40\%\) or more in generation quality (e.g., TextCaps) compared to prefill-only baselines, indicating substantially better preservation of visually grounded information required for coherent multimodal generation.

In addition to standalone performance, MM-ShiftKV remains complementary to KV budget allocation strategies. As shown in Table~\ref{tab:mm_budget_allocation_llava}, integrating MM-ShiftKV with representative budget allocation methods (e.g., PyramidKV and AdaKV) leads to consistent relative improvements, typically in the range of \(10\%\)–\(20\%\) under the most restrictive budgets. These results indicate that correcting the prefill--decode scale mismatch improves KV \emph{selection quality} independently of how KV budgets are allocated across layers or heads.

Overall, the results demonstrate that MM-ShiftKV provides a robust and effective prefill-only solution for multimodal inference under strict KV-cache constraints, offering favorable accuracy--memory trade-offs while remaining compatible with different KV budgeting schemes.

\newcommand{\cmark}{\ding{51}}  

\begin{table}[t]
\centering
\small
\setlength{\tabcolsep}{3pt}
\renewcommand{\arraystretch}{1.05}
\begin{tabular}{l c c c c c}
\toprule
Variant & Samp. & Var.~Exp. & G Vote & OCR  & TextCaps \\
\midrule
LastAttn
&        &        &        & 52.3 & 40.8 \\

+ Samp.
& \cmark &        &        & 54.5 & 45.5 \\

+ Var.~Exp.
& \cmark & \cmark &        & 59.6 & 48.6 \\

+ GVote
& \cmark & \cmark & \cmark & \textbf{68.3} & \textbf{50.4} \\
\bottomrule
\end{tabular}
\caption{
Ablation study under a fixed per-head KV cache budget.
\textbf{Samp.}, \textbf{Var.~Exp.}, and \textbf{GVote} denote query proxy sampling,
variance expansion, and group-wise voting, respectively.
}

\label{tab:ablation}
\end{table}
\subsection{Ablation Study}
\label{sec:exp:ablation}

We analyze the contribution of individual components in \textbf{MM-ShiftKV}
through ablation experiments under a fixed per-head KV cache budget.
All ablation experiments are conducted using the
\textbf{Qwen2.5-VL-7B-Instruct} model, following the same prefill-only evaluation
protocol as in the main experiments.
We start from a lightweight \emph{LastAttn} baseline, which incorporates only the
attention induced by the last prefill query, and progressively add
decode-aware query sampling, variance expansion, and group-wise voting.
Results are reported on \textbf{OCRBench} and \textbf{TextCaps} in
Table~\ref{tab:ablation}.

\paragraph{Overall Effect.}
As shown in Table~\ref{tab:ablation}, the \emph{LastAttn} baseline achieves
52.3 OCR accuracy and 40.8 CIDEr on TextCaps, providing a simple but stable
prefill-only reference.
Enabling decode-aware query sampling (\textbf{+ Samp.}) consistently improves
performance, increasing OCR accuracy to 54.5 and TextCaps CIDEr to 45.5.
This indicates that sampling query proxies aligned with the decoding stage
provides a more informative estimate of future attention behavior than relying
solely on prefilling-stage statistics.

Adding variance expansion (\textbf{+ Var.~Exp.}) yields further gains, improving
performance to 59.6 on OCRBench and 48.6 CIDEr on TextCaps.
This suggests that query proxies derived directly from prefilling statistics are
systematically under-scaled, and that scale calibration is critical for capturing
the variability of decoding-time queries in multimodal inference.

Finally, incorporating group-wise voting (\textbf{+ GVote}) achieves the best
overall results, reaching 68.3 OCR accuracy and 50.4 CIDEr on TextCaps.
By aggregating attention mass across groups of query proxies, group-wise voting
effectively reduces estimation variance and stabilizes KV ranking under tight
cache budgets.

Overall, these results demonstrate that decode-aware query sampling, variance
expansion, and group-wise voting are complementary components.
Starting from a simple last-query attention anchor, progressively introducing
decode-aware calibration and variance reduction is necessary to fully realize the performance gains of \textbf{MM-ShiftKV} in prefill-only multimodal inference.

\begin{figure}[t]
  \centering
  \includegraphics[width=\columnwidth]{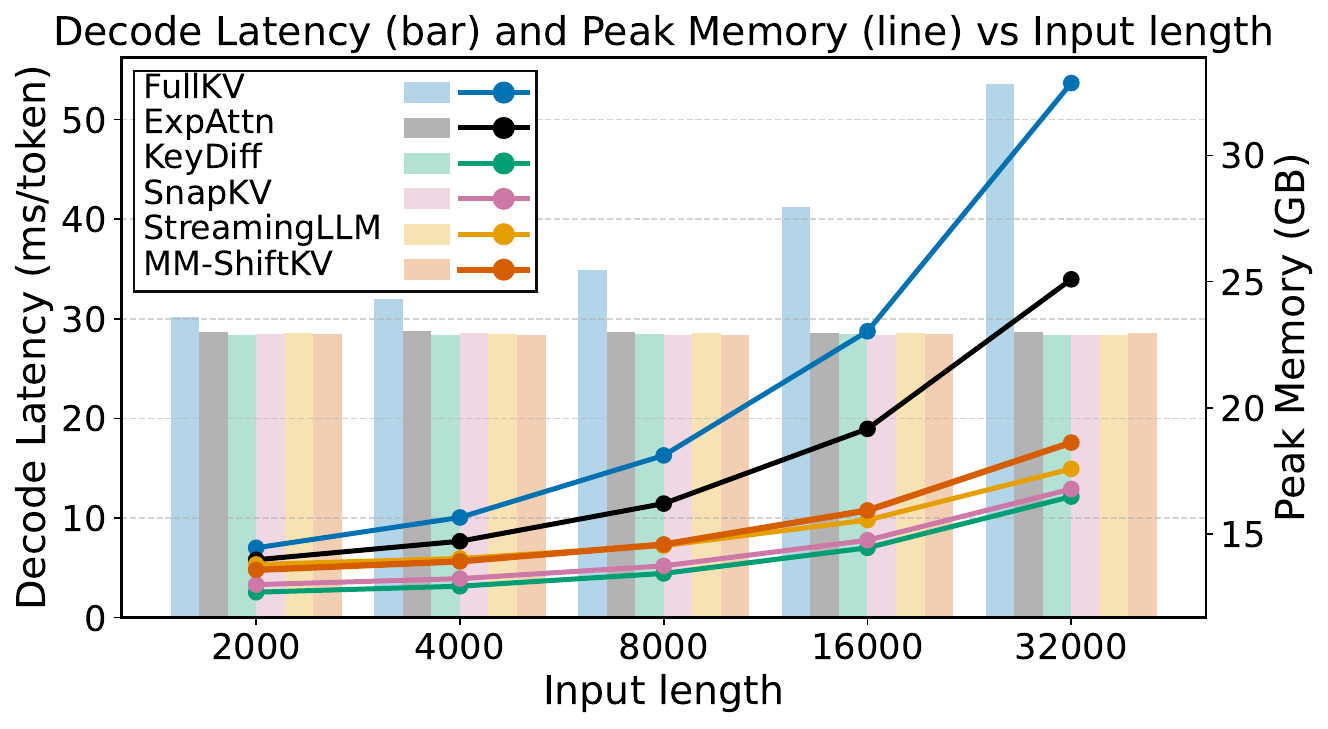}
\caption{
End-to-end decoding latency and peak GPU memory usage under increasing input
lengths.
Bars denote per-token decoding latency and lines indicate peak GPU memory usage.
All methods are evaluated with a fixed per-head KV budget of 256 and a maximum
output length of 100 tokens.
}

  \label{fig:latency_memory}
\end{figure}

\subsection{Efficiency Evaluation}
\label{sec:exp:efficiency}

\paragraph{Setup.}
We evaluate inference efficiency under long-context multimodal settings by
varying the input length in
$\{2\mathrm{K}, 4\mathrm{K}, 8\mathrm{K}, 16\mathrm{K}, 32\mathrm{K}\}$ while fixing
the output length to 100 tokens.
Following the main configuration, we apply a per-head KV cache budget of 256
after prefilling-stage compression and report per-token decoding latency and
peak GPU memory usage.
All experiments use FlashAttention-compatible decoding kernels.

\paragraph{Decoding Latency.}
As shown in Figure~\ref{fig:latency_memory}, MM-ShiftKV consistently reduces
per-token decoding latency compared to FullKV, with the gap widening as the
input length increases.
While FullKV exhibits steadily growing latency, MM-ShiftKV maintains an
almost constant latency profile, achieving up to a $1.9\times$ speedup at
32K input length.

\paragraph{Memory Cost.}
By bounding the prompt KV cache before decoding, MM-ShiftKV also substantially
reduces peak GPU memory usage.
At 32K input length, peak memory consumption is reduced from approximately
32.9~GB to 18.6~GB, corresponding to a reduction of about 43\%.
Overall, MM-ShiftKV offers a more favorable latency--memory trade-off than
existing prefilling-stage compression baselines.




\section{Related Work}
\label{sec:related}

Current research on KV cache compression can be broadly categorized into two types, static pruning~\citep{li2024snapkv,jiang2024minference,devoto2024l2kv,devoto2025expected,park2025keydiff} and dynamic eviction~\citep{chen2025fastv,child2019generating}, based on the intervention stage. Recently, hybrid KV cache compression strategies have also emerged to balance efficiency across different inference stages~\citep{zeng2026hybridkvhybridkvcache}. Static pruning methods are primarily deployed during the prefill stage, where the importance of each KV pair is evaluated using predefined metrics, and lower-scoring KV pairs are pruned to reduce the initial memory footprint. Dynamic eviction methods continuously discard tokens during the decode stage to maintain a fixed-size cache. There are three paradigms for KV cache eviction: first, attention-based eviction~\citep{li2024snapkv,devoto2025expected}; second, value-based analysis~\citep{devoto2024l2kv,park2025keydiff}; third, heuristic-based structural eviction~\citep{cai2024pyramidkv,xiao2023efficient}.

A key limitation of these methods lies in their implicit assumption of ``distributional homogeneity''~\citep{devoto2025expected,cai2024medusa}. However, MM-ShiftKV finds that in multimodal scenarios, there is a significant numerical distribution shift between these two stages. MM-ShiftKV explicitly analyzes the differences in numerical behavior between the prefill and decode stages in multimodal contexts. By introducing a group voting mechanism, MM-ShiftKV achieves more robust and accurate KV selection.

Additionally, comprehensive studies outline efficient inference bottlenecks in large vision-language models~\citep{zhang2026efficientinferencelargevisionlanguage}. Broadly, research covers quantization, early exiting, and speculative decoding~\citep{lin2024awq,elhoushi2024layerskip,child2019generating,liu2024kivi,xu2025specee,su2025rotatekv,ji2026foresttreeslooselyspeculative}, alongside hardware and system-level optimizations~\citep{dao2023flashattention2,zheng2025flashgen,zhou2025floe}. In multimodal scenarios, specialized strategies~\citep{lin2025multilayer} include dynamic attention head allocation~\citep{wang2025sparsemm,yang2025topv,wan2024lookm} and visual token pruning~\citep{shen2024longvu,yang2025visionzip,zhang2025sparsevlm}. MM-ShiftKV is complementary and seamlessly integrates with these approaches for enhanced performance.

\section{Conclusion}
\label{sec:conclusion}

We studied a previously overlooked aspect of multimodal inference: the systematic mismatch between prefill and decoding stages. By making \emph{prefill-stage} KV selection decode-aware, MM-ShiftKV provides a
simple, training-free solution that aligns prompt KV management with
decoding-time query distributions. Our results highlight decode-aware KV management as a key design principle for scalable multimodal inference.

\section{Ethical Considerations}
\label{sec:ethicalconsiderations}
All experiments in our work are conducted using open-source datasets and models. Our research is solely aimed at enabling efficient inference of multimodal large models (MLLMs), and does not involve any human subject, sensitive data, or commercial applications.

\section{Limitations}
\label{sec:limitations}
While MM-ShiftKV
 exhibits prominent advantages in decoding latency optimization and inference accuracy performance under resource-constrained conditions, it still has certain limitations: First, the additional computation introduced by the voting scoring for KV eviction processing in the prefilling phase incurs a certain overhead. Although this overhead is negligible compared to the overall inference latency, there remains room for further acceleration in the prefilling phase. Second, MM-ShiftKV performs KV eviction only in the prefilling phase. Although the KV cache in the prefilling phase accounts for the largest proportion in most visual question answering tasks, this design poses certain challenges in some tasks that require long-context reasoning. Third, our method primarily focuses on image and video benchmarks (which generate a large amount of KV cache), but it may face new challenges for other tasks and modalities (e.g., audio-based multimodal large models or long-context reasoning scenarios).

Nevertheless, MM-ShiftKV is compatible with multiple future extension directions, including combination with attention head budget allocation methods, integration of offloading techniques in the decoding phase, and joint application with other compression paradigms (e.g., model quantization, speculative decoding, etc.). We believe these directions will further enhance the generality and scalability of MM-ShiftKV.

\section*{Acknowledgments}
This work was supported by the Pioneer R\&D Program of Zhejiang Province (No.~2024C01021) and the Zhejiang Province Leading Talent of Technological Innovation Program (No.~2023R5214).

\bibliography{custom}
\FloatBarrier

\newpage
\appendix

\section{Theoretical Analysis of Prefill--Decode Scale Mismatch}
\label{sec:appendix:theory}

\begin{figure}[t]
    \centering
    \begin{subfigure}[t]{0.9\linewidth}
        \centering
        \includegraphics[width=\linewidth]{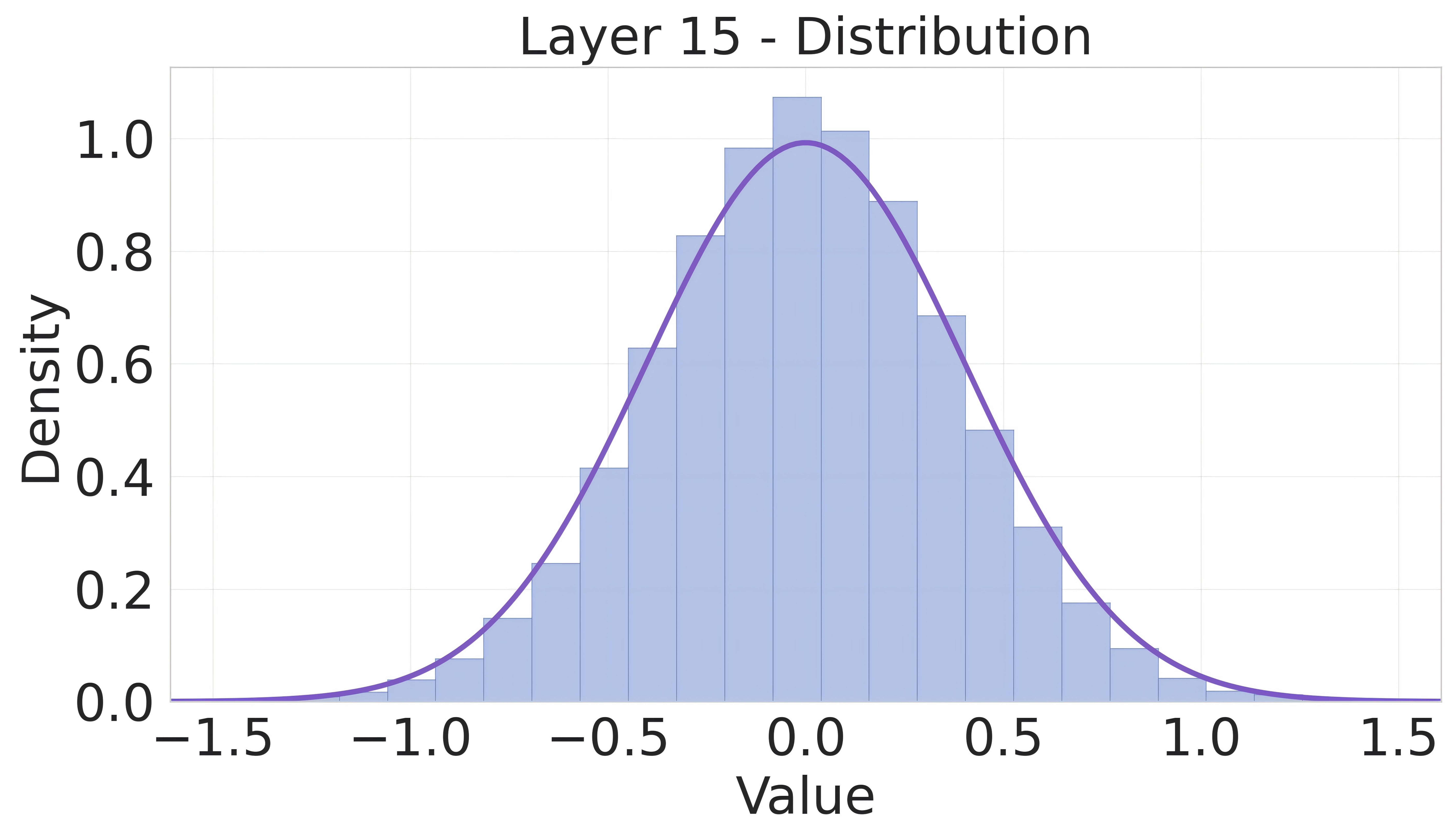}
        \caption{Layer 15}
        \label{fig:appendix:gs_layer15}
    \end{subfigure}\hfill
    \begin{subfigure}[t]{0.9\linewidth}
        \centering
        \includegraphics[width=\linewidth]{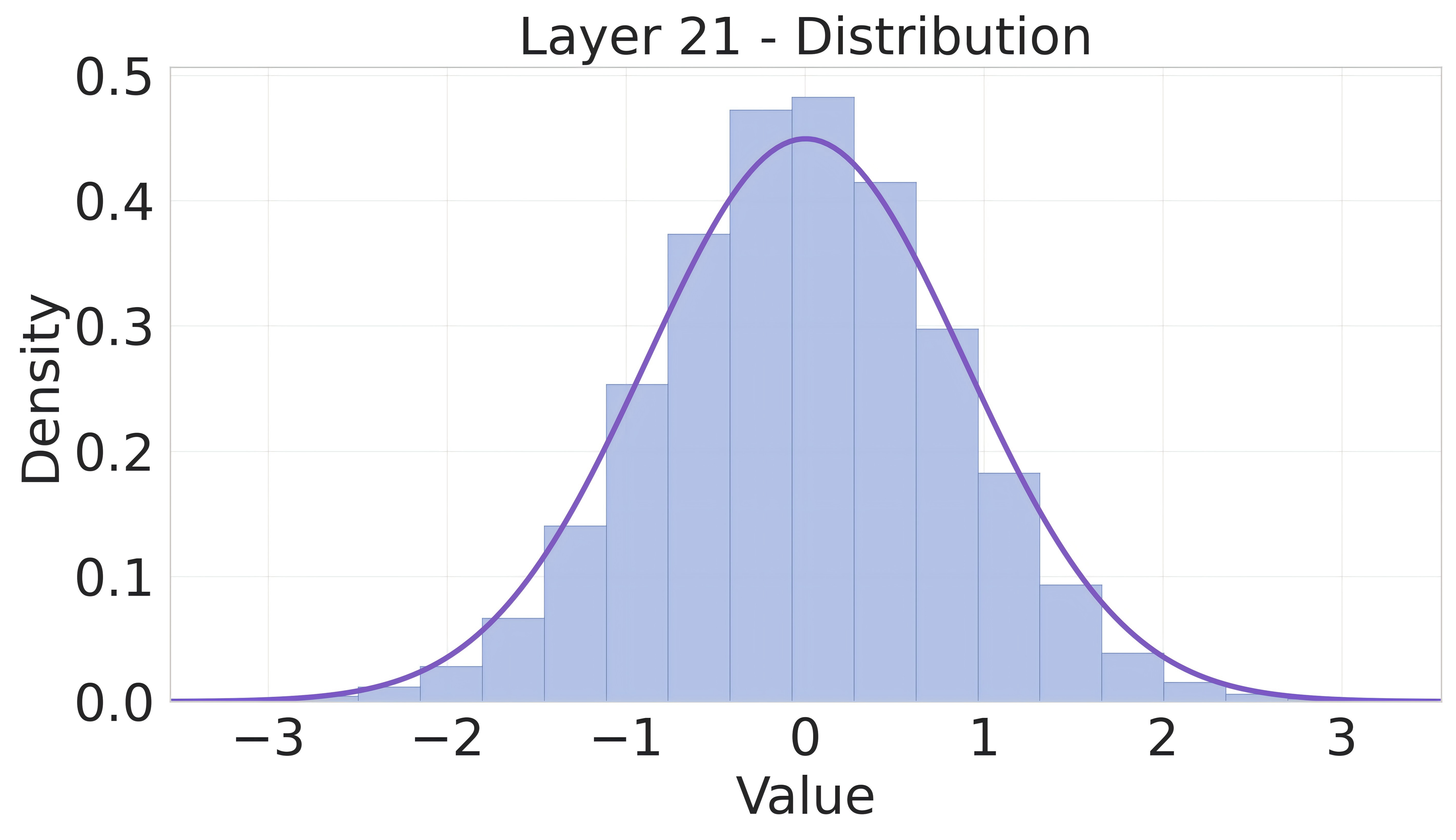}
        \caption{Layer 21}
        \label{fig:appendix:gs_layer21}
    \end{subfigure}\hfill
    \caption{The figure shows the overall numerical distribution of hidden states across different layers of the \textbf{LLaVA-NeXT-Vicuna-7B} model during inference. We flattened the hidden states to calculate their overall numerical distribution, where the x-axis represents specific values and the y-axis denotes distribution density (with the calculation formula as follows: $\mathrm{density}(x) = \frac{\mathrm{freq}(x)}{N \cdot \Delta x}$,where $\mathrm{density}(x)$: the empirical density (height of the histogram) at value $x$;$\mathrm{freq}(x)$: the number of samples falling into the interval centered at (or starting from) $x$;$N$: the total number of all samples;$\Delta x$: the width of the interval (bin).).
}
    \label{fig:appendix:theory}
\end{figure}

This appendix provides a concise theoretical explanation of the prefill--decode
scale mismatch and motivates variance-expanded, probe-based KV selection as a
principled decode-aware approximation under prefill-only execution.

\subsection{Distributional Shift Between Prefilling and Decoding}

As shown in Figure~\ref{fig:appendix:theory}, the overall numerical distribution
of hidden states across layers is approximately Gaussian-like at the projection
input. We therefore model hidden states at a given layer as Gaussian random
variables for tractable approximation.
Empirically, decoding-time hidden states and query projections exhibit larger
variance than those observed during prefilling:
\begin{equation}
\begin{array}{l}
h_{\mathrm{pre}} \sim \mathcal{N}(\mu, \Sigma_{\mathrm{pre}}), \\
h_{\mathrm{dec}} \sim \mathcal{N}(\mu, \Sigma_{\mathrm{dec}})
\end{array}
\end{equation}
where
\begin{equation}
\Sigma_{\mathrm{dec}} \succ \Sigma_{\mathrm{pre}} .
\end{equation}
Through linear query projection, this induces a corresponding variance gap in
query distributions, causing prefilling-based query proxies to underestimate
the support of true decoding-time queries.

\subsection{Impact on Attention Estimation}

For a fixed key $k$ and Gaussian query
$q \sim \mathcal{N}(\mu_q, \Sigma_q)$,
the expected unnormalized attention score admits the closed form
\begin{equation}
\mathbb{E}\!\left[
\exp\!\left(\frac{q^\top k}{\sqrt{d}}\right)
\right]
=
\exp\!\left(
\frac{\mu_q^\top k}{\sqrt{d}}
+
\frac{k^\top \Sigma_q k}{2d}
\right).
\end{equation}

Underestimating $\Sigma_q$ therefore systematically underestimates expected
attention mass, particularly for keys aligned with high-variance query
directions, leading to biased KV ranking under constrained budgets.

\subsection{First- and Second-Order Interpretation}
\label{sec:appendix:theory:alignment}

The closed-form expression above provides a direct interpretation of the two
design choices in MM-ShiftKV:
\begin{itemize}
    \item \textbf{Sample-wise mean centering preserves the first-order structure} ($\mu_q$):
    the proxy distribution is centered using the prefill statistics of the current
    input sample, which provides an input-adaptive semantic center for proxy construction.

    \item \textbf{Variance expansion compensates the second-order term} ($\Sigma_q$):
    inflating variance enlarges coverage along high-variance directions so that
    semantically important outlier-aligned keys are less likely to be underestimated.
\end{itemize}

Therefore, MM-ShiftKV does not explicitly align proxy means to decoding-stage
statistics. Instead, it preserves the sample-wise prefill mean structure for
semantic stability and corrects the dominant prefill--decode discrepancy through
variance expansion.

\subsection{Variance-Expanded and Probe-Based Approximation}

To compensate for this bias without accessing decoding-time signals, we introduce
a variance-expanded proxy distribution
\begin{equation}
\tilde{q} \sim \mathcal{N}(\mu_q, \gamma^2 \Sigma_{q,\mathrm{pre}}),
\qquad \gamma > 1 .
\end{equation}
This expansion enlarges the support of prefilling-based queries while preserving
their mean structure.

KV importance is then estimated by sampling a finite number of query probes,
yielding a Monte Carlo approximation of expected attention mass.
Group-wise aggregation further reduces estimator variance without materializing
attention matrices.

\subsection{Summary}

This analysis explains why uncalibrated prefill-based KV selection fails under
distributional scale shift and why variance-expanded, probe-based attention
estimation provides an effective and training-free decode-aware alternative for
memory-constrained multimodal inference.

\section{Implementation Details}
\label{sec:appendix:implementation}

Our method supports Grouped-Query Attention (GQA) models such as
Qwen2.5-VL-7B-Instruct and LLaVA-v1.6-Vicuna-7B.
In GQA, query states have shape $(B, L, H_q, d)$ and key-value states stored
in the KV cache have shape $(B, L, H_{kv}, d)$, with $H_q = H_{kv} \times G$.
For attention computation, we repeat the key and value states along the
head dimension to restore an MHA-like layout, yielding attention scores of
shape $(B, H_q, L_q, L_k)$.
Although attention scores are computed at the query-head level, KV cache
budgeting is performed at the key-value head level by aggregating the
scores of query heads belonging to the same key-value head.
KV importance is estimated during prefilling using synthetic query probes
sampled from statistics computed over the full prefill sequence, with $N=512$ samples per
layer, generated from a diagonal Gaussian distribution parameterized by sample-wise
prefill statistics.
The probe queries are grouped into $G_{\mathrm{groups}}=32$ groups to
stabilize estimation, and tokens are ranked by aggregated attention scores.
For each key-value head, we select the smallest set of tokens whose
cumulative attention mass exceeds a fixed threshold (0.95), while always
preserving the most recent token and incorporating the attention score
from the last real query token.
The selected tokens are restored to their original temporal order,
concatenated with the most recent token, and inserted into the KV cache
using the standard update interface, ensuring full compatibility with
FlashAttention-style kernels.
All experiments are conducted in a purely inference-time setting without
additional training or fine-tuning.

\begin{algorithm}[t]
\caption{MM-ShiftKV (Prefill-only, Decode-aware KV Selection)}
\label{alg:mm-shiftkv}
\small
\begin{algorithmic}[1]
\Require
Prefill hidden states $\{h_t^{(\ell)}\}_{t=1}^{T}$,
prompt KVs $\{(k_t^{(\ell,h)},v_t^{(\ell,h)})\}_{t=1}^{T}$,
budget $C_{\ell,h}$
\Ensure
Compressed KV cache $\mathcal{C'}^{(\ell,h)}$

\State Compute prefill statistics
$(\mu_{\text{pre}}^{(\ell)}, \sigma_{\text{pre}}^{(\ell)})$
\State Sample $N{=}Gg$ hidden states with variance expansion $\gamma$
\State Project samples to query proxies and apply future-position RoPE
\State Partition query proxies into $G$ groups of size $g$

\For{each group $g'$}
  \State Aggregate attention mass over prompt keys
  \State Select minimal set covering fraction $\tau$
  \State Vote selected tokens
\EndFor

\State Add last-query anchor:
$s_t \leftarrow \mathrm{vote}(t) + \lambda a_t(q_{\text{last}}^{(\ell,h)})$
\State Always retain token $T$ and select top-$C_{\ell,h}-1$ tokens by $s_t$
\State Restore temporal order and return $\mathcal{C'}^{(\ell,h)}$
\end{algorithmic}
\end{algorithm}


\begin{figure}[t]
    \centering
    \begin{subfigure}[t]{0.85\linewidth}
        \centering
        \includegraphics[width=\linewidth]{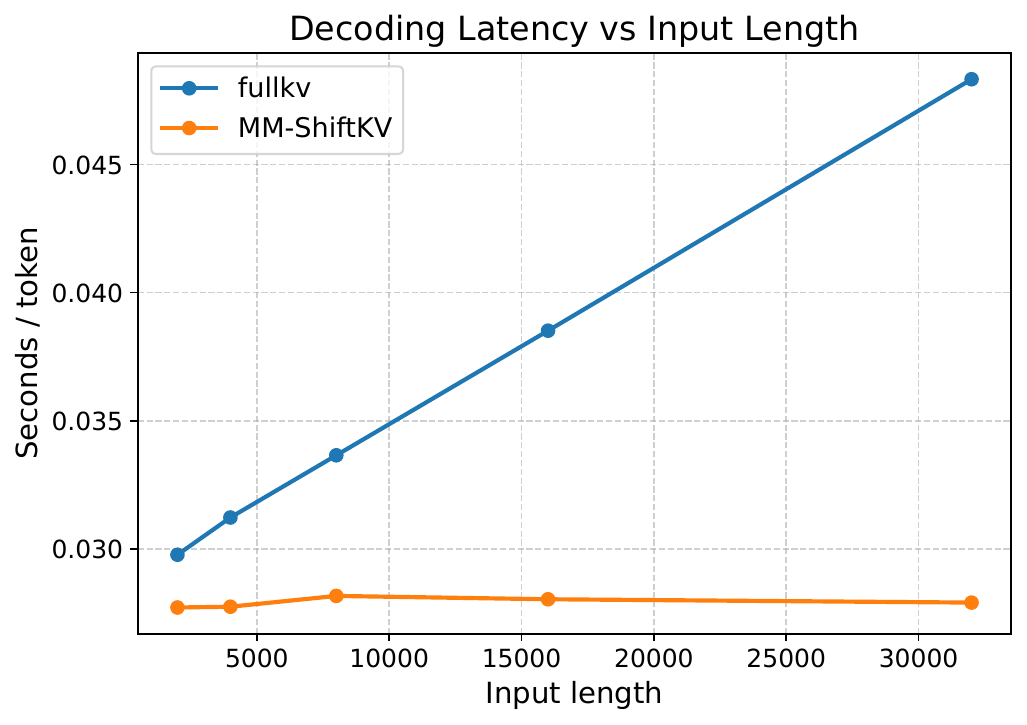}
        \caption{Per-token decoding latency vs.\ input length.}
        \label{fig:appendix:decoding_latency}
    \end{subfigure}
    \hfill
    \begin{subfigure}[t]{0.85\linewidth}
        \centering
        \includegraphics[width=\linewidth]{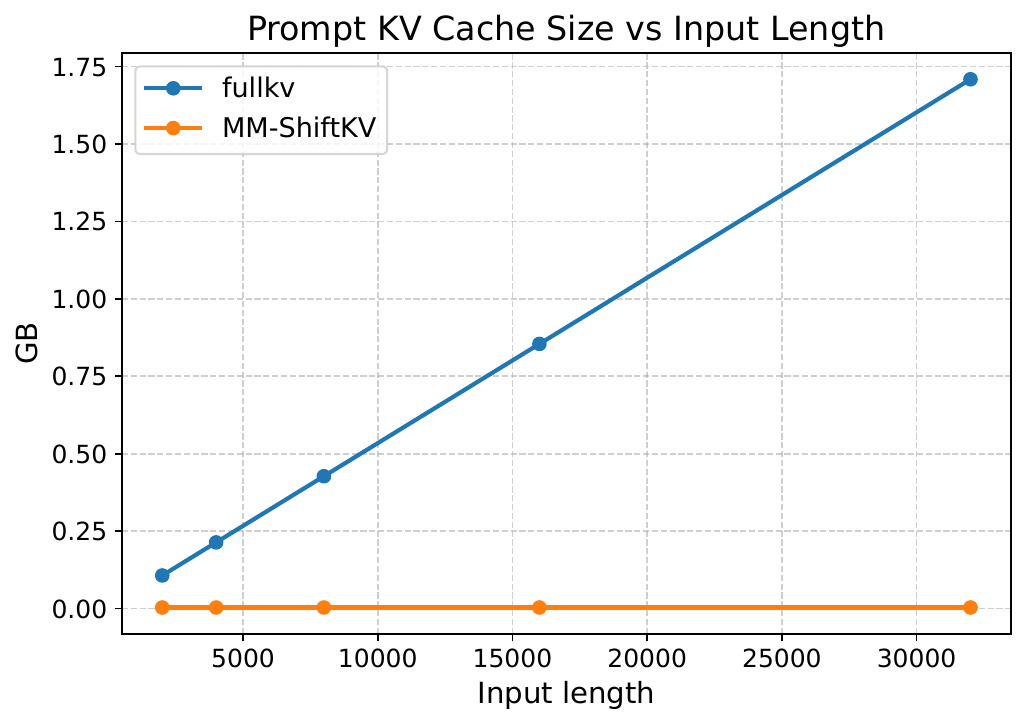}
        \caption{Prompt KV cache size vs.\ input length.}
        \label{fig:appendix:cache_size}
    \end{subfigure}

    \caption{
        Visualization of decoding latency and prompt KV cache size under increasing
        input lengths.
    }
    \label{fig:appendix:efficiency}
\end{figure}

\begin{figure}[t]
    \centering
    \begin{subfigure}[t]{0.95\linewidth}
        \centering
        \includegraphics[width=\linewidth]{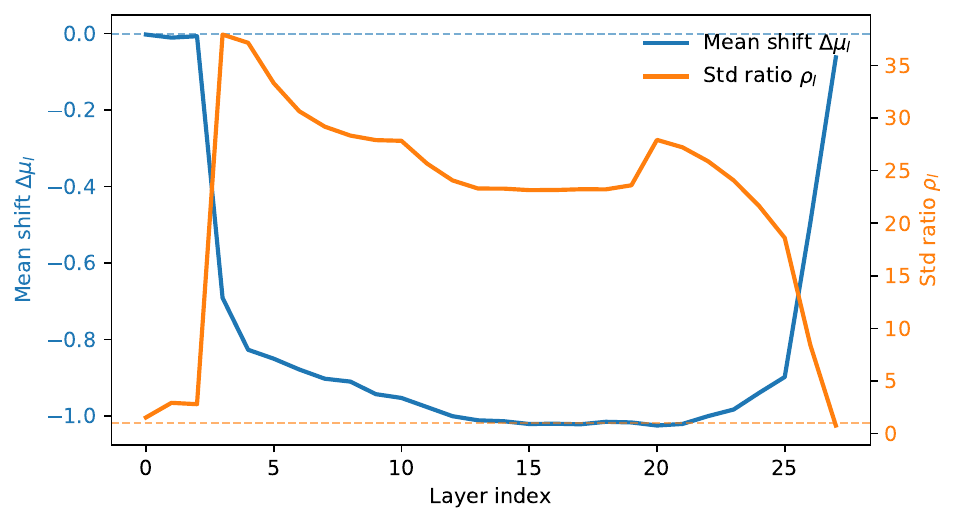}
        \caption{DocVQA}
        \label{fig:appendix:docvqa_stats}
    \end{subfigure}
    \vspace{0.4em}
    \begin{subfigure}[t]{0.95\linewidth}
        \centering
        \includegraphics[width=\linewidth]{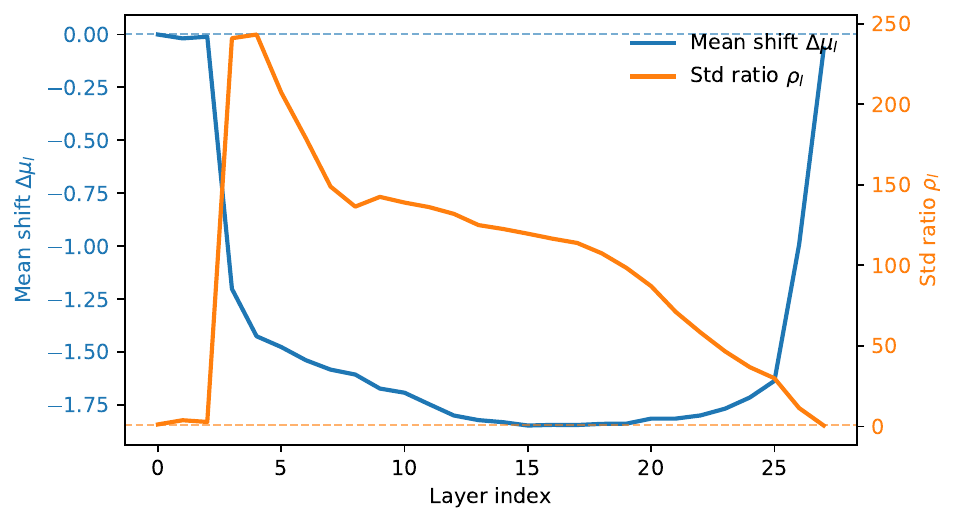}
        \caption{SynthDog}
        \label{fig:appendix:synthdog_stats}
    \end{subfigure}

    \caption{
    Additional visualizations of the prefill--decode \emph{scale mismatch}
    in multimodal inference across different datasets.
    \textbf{(a)} Layer-wise representation statistics on \textbf{DocVQA},
    showing that decoding-stage hidden states exhibit substantially larger
    variance than those observed during prefilling.
    \textbf{(b)} The same prefill--decode variance mismatch observed on
    \textbf{SynthDog}, indicating that the scale mismatch is consistent
    across datasets.
    }
    \label{fig:appendix:visual_analysis}
\end{figure}

\section{Additional Visualization Results}
\label{sec:appendix:visual}

We present additional visualizations to complement the statistical observations
in Section~\ref{sec:observation} and to provide intuitive evidence of the
prefill--decode \emph{scale mismatch} in long-context multimodal inference.
All visualizations are obtained using \textbf{Qwen2.5-VL-7B-Instruct} on representative
document understanding benchmarks, including \textbf{DocVQA} and
\textbf{SynthDog}.

As shown in Figure~\ref{fig:appendix:efficiency}, FullKV exhibits steadily
increasing per-token decoding latency and prompt KV cache size as the input
length grows.
In contrast, MM-ShiftKV bounds the prompt KV cache after prefilling, resulting
in near-constant decoding latency and significantly reduced memory usage.
This visualization illustrates how prefill-only KV selection decouples decoding
cost from the original input length under long-context multimodal inputs.

Figure~\ref{fig:appendix:visual_analysis} provides additional evidence of the
prefill--decode \emph{representation scale mismatch} observed in
Section~\ref{sec:observation}.
Specifically, Figure~\ref{fig:appendix:docvqa_stats} visualizes layer-wise
representation statistics on \textbf{DocVQA}, showing that decoding-stage hidden
states exhibit consistently larger variance than those observed during
prefilling, despite sharing identical model parameters.

Figure~\ref{fig:appendix:synthdog_stats} shows that the same variance expansion
effect persists on \textbf{SynthDog}, indicating that the prefill--decode scale
mismatch is not specific to a single dataset but instead reflects a systematic
property of multimodal inference under long contexts.
Together, these visualizations demonstrate that prefilling-stage statistics
systematically underestimate the scale of decoding-time representations across
different document understanding benchmarks.

\section{Sensitivity Study on Hyperparameters}
\label{sec:appendix:sensitivity}

We conduct a sensitivity study to examine the robustness of \textbf{MM-ShiftKV}
with respect to its key hyperparameters.
Unless otherwise specified, all experiments in this section are performed on
\textbf{Qwen2.5-VL-7B-Instruct} under a fixed per-head KV cache budget of
$C{=}64$.
We report results on three representative multimodal benchmarks:
\textbf{OCRBench} and \textbf{TextVQA} (accuracy), and \textbf{TextCaps} (CIDEr).
When analyzing one hyperparameter, all others are held fixed at their default
values ($\gamma{=}10$, $N{=}512$, $G{=}32$, $\tau{=}0.95$, $\lambda{=}1$).

\subsection{Variance Expansion Factor $\gamma$}
\label{sec:appendix:sensitivity:gamma}

The variance expansion factor $\gamma$ controls the scale calibration between
prefilling-stage query proxies and decoding-time query distributions.
As discussed in Section~\ref{sec:observation}, decoding-time queries exhibit
substantially larger variance than those observed during prefilling, motivating
the use of $\gamma>1$.

Table~\ref{tab:sensitivity_gamma} shows that setting $\gamma{=}1$, corresponding
to no variance expansion, leads to consistently degraded performance across all
benchmarks.
Increasing $\gamma$ significantly improves performance, indicating that scale
calibration is critical for effective KV importance estimation.
Performance peaks around $\gamma{=}10$, while further increasing $\gamma$ yields
diminishing returns.
Based on this observation, we fix $\gamma{=}10$ as the default value in all
experiments.

\subsection{Attention Mass Threshold $\tau$}
\label{sec:appendix:sensitivity:tau}

The attention mass threshold $\tau$ determines the minimum cumulative attention
mass preserved when selecting prompt KV tokens.
A smaller threshold may discard occasionally important tokens, whereas an overly
large threshold approaches FullKV behavior and weakens compression.

As shown in Table~\ref{tab:sensitivity_tau}, $\tau{=}0.95$ consistently achieves
the best trade-off between performance and compression across all evaluated
benchmarks.
Lower thresholds result in noticeable performance drops, while higher thresholds
provide limited additional benefit.
We therefore adopt $\tau{=}0.95$ as a stable default setting.

\subsection{Number of Query Proxies $N$}
\label{sec:appendix:sensitivity:N}

The number of query proxies $N$ controls the quality of the Monte Carlo
approximation of expected attention mass.
Larger $N$ reduces estimator variance but increases prefilling-stage computation.

Results in Table~\ref{tab:sensitivity_N} show that performance improves as $N$
increases from 128 to 512 and stabilizes at $N{=}512$.
Using fewer proxies leads to noisier importance estimates, while larger values
offer limited additional gains relative to the increased overhead.
We therefore set $N{=}512$ as a balanced choice between accuracy and efficiency.

\subsection{Summary}
\label{sec:appendix:sensitivity:summary}

Overall, these sensitivity studies demonstrate that \textbf{MM-ShiftKV} is robust
to moderate variations in its hyperparameters.
The selected default values correspond to stable operating points that
consistently balance accuracy, memory efficiency, and prefilling-stage overhead
across OCR-centric, multimodal question answering, and image-conditioned
generation tasks.

\begin{table}[t]
\centering
\small
\setlength{\tabcolsep}{6pt}
\renewcommand{\arraystretch}{1.05}
\begin{tabular}{c c c c c}
\toprule
$\gamma$
& OCRBench
& TextVQA
& TextCaps
& Avg \\
\midrule
1  & 55.7 & 66.5 & 51.8 & 58.0 \\
2  & 56.1 & 69.2 & 55.1 & 60.1 \\
5  & 65.4 & 73.5 & 58.6 & 65.8 \\
10 & \textbf{68.8} & \textbf{80.0} & \textbf{63.2} & \textbf{70.7} \\
20 & 68.5 & 79.3 & 60.1 & 69.3 \\
\bottomrule
\end{tabular}
\caption{
Sensitivity study on the variance expansion factor $\gamma$.
OCRBench and TextVQA are evaluated with accuracy, and TextCaps with CIDEr.
Avg denotes the macro-average over the three benchmarks.
The per-head KV cache budget is $C{=}64$.
Higher is better.
}
\label{tab:sensitivity_gamma}
\end{table}




\begin{table}[t]
\centering
\small
\setlength{\tabcolsep}{6pt}
\renewcommand{\arraystretch}{1.05}
\begin{tabular}{c c c c c}
\toprule
$\tau$
& OCRBench
& TextVQA
& TextCaps
& Avg \\
\midrule
0.90 & 62.4 & 78.1 & 42.4 & 61.0 \\
0.95 & \textbf{68.8} & \textbf{80.0} & \textbf{63.2} & \textbf{70.7} \\
0.99 & 66.7 & 79.2 & 49.0 & 65.0 \\
\bottomrule
\end{tabular}
\caption{
Sensitivity study on the attention mass threshold $\tau$.
Metrics and averaging follow Table~\ref{tab:sensitivity_gamma}.
The per-head KV cache budget is $C{=}64$.
}
\label{tab:sensitivity_tau}
\end{table}



\begin{table}[t]
\centering
\small
\setlength{\tabcolsep}{6pt}
\renewcommand{\arraystretch}{1.05}
\begin{tabular}{c c c c c}
\toprule
$N$
& OCRBench
& TextVQA
& TextCaps
& Avg \\
\midrule
128 & 63.1 & 78.6 & 49.8 & 63.8 \\
256 & 63.5 & 78.9 & 47.0 & 63.1 \\
512 & \textbf{68.8} & \textbf{80.0} & \textbf{63.2} & \textbf{70.7} \\
\bottomrule
\end{tabular}
\caption{
Sensitivity study on the number of query probes $N$.
Metrics and averaging follow Table~\ref{tab:sensitivity_gamma}.
The per-head KV cache budget is $C{=}64$.
}
\label{tab:sensitivity_N}
\end{table}


\section{Additional Experiments}
\label{sec:appendix:rebuttal}

To provide a deeper understanding of the underlying mechanisms of MM-ShiftKV, this appendix presents additional empirical evaluations. We focus our analysis on three critical dimensions: the end-to-end efficiency trade-off, the statistical robustness of our global hyperparameter strategy, and the dynamic decode-awareness under shifting context lengths.

\subsection{End-to-End Latency and Memory Breakdown}
\label{sec:appendix:rebuttal:latency}

Table~\ref{tab:rebuttal_latency} reports TTFT, end-to-end latency, and peak VRAM on a 32K-token setting with \textbf{LLaVA-v1.6-7B}. While MM-ShiftKV introduces a moderate TTFT increase (approximately 0.38s) due to the one-time proxy sampling in the prefill stage, this overhead accounts for only $\sim$2.2\% of the total inference time. This marginal initialization cost is heavily outweighed by a 42\% reduction in total end-to-end latency and a 43.3\% decrease in peak VRAM compared to FullKV. Since the proxy sampling complexity remains linear $\mathcal{O}(N \cdot K)$ with respect to sequence length, the method provides a highly cost-effective trade-off for latency-sensitive, long-context applications without introducing any decoding-time overhead.

\begin{table}[t]
\centering
\footnotesize
\setlength{\tabcolsep}{3pt}
\renewcommand{\arraystretch}{0.95}
\begin{tabular}{lccc}
\toprule
Method & TTFT [s] & Total [s] & Peak VRAM [GB] \\
\midrule
FullKV & 1.662 & 29.26 & 32.87 \\
SnapKV & 1.664 & 16.58 & 17.28 \\
Expected Attention & 1.833 & 16.92 & 25.09 \\
MM-ShiftKV & 2.042 & 16.98 & 18.62 \\
\bottomrule
\end{tabular}
\caption{
Latency and memory summary on a 32K-token setting.
TTFT denotes Time-To-First-Token.
Lower is better for all metrics.
}
\label{tab:rebuttal_latency}
\end{table}

\subsection{Global vs.\ Layer-Wise Expansion Strategies}
\label{sec:appendix:rebuttal:global_vs_layerwise}

Table~\ref{tab:rebuttal_global_layerwise} compares our globally unified strategy against a refined layer-wise variant. Empirical analysis across diverse architectures (e.g., MHA in LLaVA and GQA in Qwen2.5-VL) reveals that variance expansion is an intrinsic property of MLLM inference, consistently expanding by an order of magnitude (typically $13\times$ to $22\times$) from prefill to decoding. While actual variance fluctuates across layers, fine-grained layer-wise tuning often underestimates the required bandwidth for layers with smaller prefill deviations, leading to diversity collapse. In contrast, a globally fixed $\gamma$ acts as a robust safety upper-bound and broad search bandwidth, effectively preventing the loss of critical outliers without the need for ad-hoc per-model calibration.

\begin{table}[t]
\centering
\footnotesize
\setlength{\tabcolsep}{4pt}
\renewcommand{\arraystretch}{0.95}
\begin{tabular}{lcc}
\toprule
Strategy & OCRBench & Avg \\
\midrule
Layer-wise expansion & 66.2 & 65.9 \\
Globally unified (ours) & \textbf{68.8} & \textbf{66.9} \\
\bottomrule
\end{tabular}
\caption{
Comparison of global and layer-wise expansion strategies on Qwen2.5-VL-7B.
Higher is better.
}
\label{tab:rebuttal_global_layerwise}
\end{table}

\subsection{Mean-Centering Strategy Ablation}
\label{sec:appendix:rebuttal:mean_centering}

Table~\ref{tab:rebuttal_schedule} compares different mean-centering strategies
for query proxy construction. Since MM-ShiftKV does not explicitly align proxy
means to decoding-stage statistics, we instead study how different prefill-based
centering schemes affect robustness. In particular, we compare no mean centering,
a globally shared prefill mean, and sample-wise mean centering computed from the
current input. Sample-wise mean centering performs best, indicating that
adapting the proxy center to each input sample provides a better semantic anchor
than a fixed global center. The main gain of MM-ShiftKV still comes from variance
expansion, while sample-wise centering further improves stability.

\begin{table}[t]
\centering
\footnotesize
\setlength{\tabcolsep}{4pt}
\renewcommand{\arraystretch}{0.95}
\begin{tabular}{lc}
\toprule
Mean-Centering Strategy & OCRBench (Acc.) \\
\midrule
No mean centering & 58.4 \\
Global mean centering & 65.5 \\
Sample-wise mean centering (ours) & \textbf{68.8} \\
\bottomrule
\end{tabular}
\caption{
Ablation on mean-centering strategies for query proxy construction
(Qwen2.5-VL-7B, OCRBench).
Higher is better.
}
\label{tab:rebuttal_schedule}
\end{table}

\subsection{Sensitivity to Anchor Weight $\lambda$}
\label{sec:appendix:rebuttal:lambda}

To clarify the role of the last-query anchor, we vary $\lambda$ while keeping all other settings fixed (Qwen2.5-VL-7B, OCRBench, $C{=}64$). MM-ShiftKV utilizes a lexicographical-like hierarchical ranking mechanism rather than simple linear weighting. Since proxy votes are discrete integers, setting $\lambda{=}1$ ensures the continuous anchor score acts strictly as a secondary tie-breaker. As shown in Table~\ref{tab:rebuttal_lambda}, removing the anchor ($\lambda{=}0$) forces unstable random dropping for tied tokens, degrading accuracy. Conversely, an oversized weight ($\lambda{=}50$) allows local heuristics to override the global proxy consensus, weakening the proxy-vote dominance and leading to suboptimal context retention.

\begin{table}[t]
\centering
\footnotesize
\setlength{\tabcolsep}{4pt}
\renewcommand{\arraystretch}{0.95}
\begin{tabular}{lc}
\toprule
$\lambda$ setting & OCRBench (Acc.) \\
\midrule
0 (no anchor) & 62.0 \\
1 (default) & \textbf{68.8} \\
50 (oversized) & 64.8 \\
\bottomrule
\end{tabular}
\caption{
Sensitivity analysis of last-query anchor weight $\lambda$
(Qwen2.5-VL-7B, OCRBench, budget $C{=}64$).
Higher is better.
}
\label{tab:rebuttal_lambda}
\end{table}

\subsection{Unified vs.\ Modality-Specific Processing}
\label{sec:appendix:rebuttal:modality}

We further compare unified and modality-specific statistic construction on OCRBench (Qwen2.5-VL-7B, $C{=}64$). In multimodal inference, text prompts serve as crucial anchors that guide the model to specific visual regions, exhibiting strong cross-modal coupling. Decoupling textual and visual modalities into disjoint proxy spaces disrupts unified attention scaling and fragments the evaluation pool. Table~\ref{tab:rebuttal_modality} confirms that unified processing preserves this critical cross-modal alignment, significantly outperforming modality-separated strategies and demonstrating the necessity of joint cross-modal statistics for KV selection.

\begin{table}[t]
\centering
\footnotesize
\setlength{\tabcolsep}{4pt}
\renewcommand{\arraystretch}{0.95}
\begin{tabular}{lc}
\toprule
Processing Strategy & OCRBench (Acc.) \\
\midrule
Vision-only & 32.1 \\
Text-only & 30.7 \\
Text-Vision (separate stats) & 39.8 \\
All (unified, ours) & \textbf{41.1} \\
\bottomrule
\end{tabular}
\caption{
Unified vs.\ modality-specific processing under the same KV budget
(Qwen2.5-VL-7B, OCRBench, $C{=}64$).
Higher is better.
}
\label{tab:rebuttal_modality}
\end{table}

\subsection{Distribution Shape Check at Sampling Entrance}
\label{sec:appendix:rebuttal:gaussian_like}

To clarify the Gaussian modeling assumption, MM-ShiftKV does not assume all internal activations are strictly Gaussian. Instead, it accurately models the ``layered distributional evolution.'' Due to Central Limit Theorem (CLT)-like residual aggregation, accumulated hidden states at the \emph{sampling entrance} (before query projection) maintain a highly stable, unimodal Gaussian-like profile. As shown in Figure~\ref{fig:appendix:theory}, this approximation is stable enough for proxy construction, providing a robust mathematical base. The actual heavy-tailed outliers emerge downstream during the projection phase, where weight matrices act as feature amplifiers. Our method samples at this statistically stable entry point and relies on variance expansion to bridge the subsequent structural amplification.

\subsection{Outlier Coverage Under Variance Expansion}
\label{sec:appendix:rebuttal:outlier}

To directly address the heavy-tail concern, we measure whether sampled proxies can cover critical outlier channels after projection. Following the rebuttal protocol, we track Top-0.1\% systematic outlier dimensions. Under narrow sampling ($\gamma{=}1$), probes fail to perceive outliers at the long tails due to rapid normal distribution decay, resulting in ``information blindness'' (12.4\% hit rate). As shown in Table~\ref{tab:rebuttal_outlier}, explicitly inflating the covariance via variance expansion mathematically fattens the sampling envelope, surging the outlier hit rate to 89.7\% and substantially improving coverage across heavy-tailed directions.

\begin{table}[t]
\centering
\footnotesize
\setlength{\tabcolsep}{4pt}
\renewcommand{\arraystretch}{0.95}
\begin{tabular}{lccc}
\toprule
Sampling Strategy & $\gamma$ & Outlier Hit Rate & OCRBench \\
\midrule
Narrow distribution & 1 & 12.4\% & 52.4 \\
MM-ShiftKV (ours) & 10 & 89.7\% & \textbf{68.8} \\
\bottomrule
\end{tabular}
\caption{
Coverage of critical outlier channels under different sampling scales
(Qwen2.5-VL-7B, budget $C{=}64$).
Higher is better.
}
\label{tab:rebuttal_outlier}
\end{table}
\section{Details of Baselines and Dataset}
\label{sec:appendix:datails}
This appendix presents the implementation details and specific parameter configurations of the baselines, along with the detailed content and tasks of the datasets.

\subsection{Details of Baselines}
\label{sec:appendix:detailsbase}
For our experiments, we use four methods, namely StreamingLLM, SnapKV, KeyDiff, and ExceptAttn, as our test baselines. We also compare the performance differences between all these methods and FullKV under budget-constrained conditions.

StreamingLLM is a classic heuristic KV eviction method. It introduces the concept of attention sink, retains the initial KV pairs statically, and leverages a sliding window mechanism to continuously preserve the KV pairs within the most recent window during the decoding stage. This method discards a large number of redundant intermediate KV pairs. It achieves favorable performance in text reasoning due to its streaming inference paradigm. But in multimodal scenarios, it discards a large number of critical visual KV pairs, leading to performance collapse. In the experiments on StreamingLLM, we adopt the optimal parameters specified in its original paper: to retain $4$ attention sinks, set the window size to $ \text{budgets-4} $, and to ensure fairness of comparative experiments, we do not perform KV eviction during the decoding stage.

SnapKV is a strong baseline method for large language models. It compares the similarity between the query attention scores of the final window in the prefilling stage and those in the decoding stage. It uses the final window of the prefilling stage to score the prefix KV pairs based on attention, selects the KV pairs with high attention scores, and statically retains the final window to maintain the characteristics of streaming inference. Following the optimal parameters provided by SnapKV, we set the window size to {32}, the convolution size to {5}, and the inter-group pooling to average pooling. SnapKV achieves outstanding performance on large language models and also has certain generality in multimodal scenarios, but it performs poorly under ultra-low budget conditions. We conduct a theoretical analysis of this issue: the statically retained window in SnapKV involves some waste and fails to truly evaluate the actually required queries, while reducing the window size leads to instability in KV selection.

KeyDiff achieves state-of-the-art metrics on large language models. It evaluates the relationship between the cosine similarity of Keys in large language models and the magnitude of the attention scores they receive. It proposes a query-agnostic method that scores Keys based on the cosine similarity between them, and the KV pairs with low cosine similarity are retained. In this experiment, since our dataset features streaming inference, we statically retain the last 1 token in accordance with KeyDiff’s handling of streaming inference and for the fairness of the experiment. KeyDiff is also perturbed in multimodal scenarios. The relationship between the cosine similarity of Keys for tokens in multimodal scenarios and the attention scores is inconsistent with that in text-only unimodal scenarios, which is the main reason for the decrease in KeyDiff’s accuracy in multimodal scenarios.

ExceptAttn also leverages the regularity of numerical distributions, but it assumes the distribution homogeneity between the prefilling and decoding stages. However, in multimodal scenarios, there is a certain deviation between the distributions of these two stages, which results in relatively low performance. In this experiment, we set its sampling count to 512 to ensure fairness, and we also configure the static retention of 4 attention sinks, consistent with the setup in its original paper.

\subsection{Details of Dataset}
\label{sec:appendix:detailsdata}
To comprehensively evaluate MM-ShiftKV
, we utilize the lmms-eval~\citep{zhang2024lmms_eval} evaluation framework. We employ OCRbench to assess OCR tasks and cross-modal text understanding tasks: this dataset covers multi-scenario images such as document scans and street view recognition, and we adopt accuracy as the corresponding evaluation metric. TextVQA, which spans real-world scenarios including restaurant menus and street signs, is used to evaluate image-text information question answering tasks, with performance measured by the Exact Match ~\citep{elmaalouly2022exact} metric. Each image in TextCaps is annotated with 3-5 reference descriptions that contain key text; we leverage this dataset to assess semantically consistent text-image captioning tasks, with CIDEr~\citep{vedantam2014cider} serving as the performance metric. ChartQA primarily composed of numerical charts is used to evaluate chart data understanding and reasoning-based question answering tasks, with performance assessed using relative error and the Exact Match metric. DocVQA covers a large volume of document images, and we use the \textbf{ANLS}~\citep{peer2024anls} to evaluate MM-ShiftKV’s performance on document image question answering tasks. MMMU is a dataset for calculation, geometric proof, and logical reasoning tasks. We use it to evaluate MM-ShiftKV’s impact on reasoning capabilities and whether it generates hallucinations, with accuracy as the evaluation metric.

\section{Case Study}
\label{sec:appendix:case_study}

\begin{figure*}[t]
    \centering
    \includegraphics[width=0.9\linewidth]{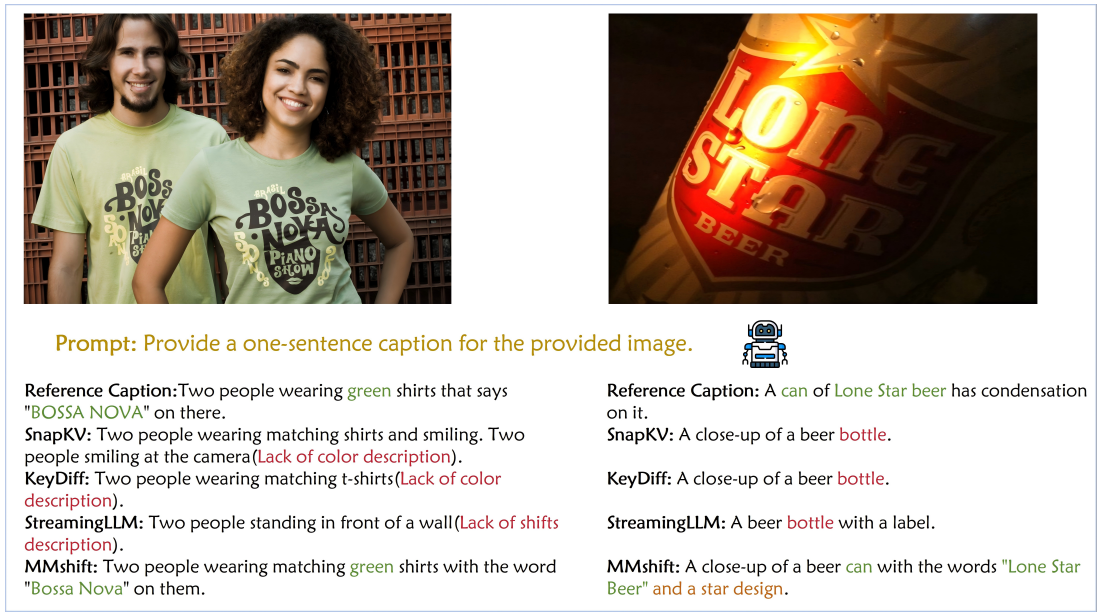}
    \caption{In the left example, the image contains two people wearing \textbf{green shirts} with the printed text \textbf{``Bossa Nova''}. While the full KV model produces an accurate caption, several baselines degrade significantly after KV compression. \textbf{SnapKV} and \textbf{KeyDiff} identify the two people but omit the color attribute, and \textbf{StreamingLLM} further loses the shirt-related information. In contrast, \textbf{MM-ShiftKV} preserves both the color and the textual content, yielding a caption consistent with the reference.In the right example, the image shows a \textbf{Lone Star Beer can} with visible branding.Under the same tight per-head budget, \textbf{SnapKV} and \textbf{KeyDiff} misclassify the object as a beer bottle or miss the embedded text, whereas \textbf{MM-ShiftKV} correctly captures both the object type and textual content. These examples demonstrate that decode-aware KV selection enables MM-ShiftKV to remain robust even under extreme KV-cache compression, consistent with its quantitative gains on TextCaps.}
    \label{fig:appendix:cs}
\end{figure*}

We present a qualitative case study on the \textbf{TextCaps} dataset using
\textbf{LLaVA-v1.6-Vicuna-7B} under an \textbf{extreme KV-cache budget of 64 tokens
per KV head}, to illustrate the behavior of different prefill-only KV selection
methods under severe memory constraints.
Figure~\ref{fig:appendix:cs} shows two representative examples requiring accurate
recognition of visual attributes and embedded text.Red denotes missing information, green indicates that key information is captured, and orange signifies that more information is captured compared to FullKV.

\end{document}